\documentclass[10pt,twocolumn,letterpaper]{article}

\usepackage{iccv}              

\usepackage{fancyhdr}
\usepackage{fancyhdr}
\fancypagestyle{firstpage}{
  \fancyhf{}
  \fancyhead[L]{\footnotesize Accepted at ICCV 2025}
  \fancyfoot[C]{\thepage}
}

\definecolor{iccvblue}{rgb}{0.21,0.49,0.74}
\usepackage[pagebackref,breaklinks,colorlinks,allcolors=iccvblue]{hyperref}

\usepackage{booktabs,tabularx}

\def\paperID{4} 
\def\confName{ICCV}
\def\confYear{2025}

\title{Who's Asking? Investigating Bias Through the Lens of Disability-Framed Queries in LLMs}

\author{%
  Vishnu Hari\textsuperscript{1}\thanks{Indicates equal contribution.}\quad
  Kalpana Panda\textsuperscript{1}\footnotemark[1]\quad
  Srikant Panda\textsuperscript{2}\quad
  Amit Agarwal\textsuperscript{2}\quad
  Hitesh Laxmichand Patel\textsuperscript{2}\\[0.4em]
  \textsuperscript{1}Birla Institute of Technology and Science (BITS) \qquad
    \textsuperscript{2}Oracle AI\\[0.3em]
  \\
  {\tt\small\{f20220094,f20220001\}@pilani.bits-pilani.ac.in}\\
  {\tt\small\{srikant.p.panda,amit.h.agarwal,hitesh.laxmichand.patel\}@oracle.com}
}

\begin{document}

\maketitle
\thispagestyle{firstpage}
\begin{abstract}
Large Language Models (LLMs) routinely infer users’ demographic traits from phrasing alone, which can result in biased responses, even when no explicit demographic information is provided. The role of disability cues in shaping these inferences remains largely uncharted. Thus, we present the first systematic audit of disability‑conditioned demographic bias across eight state‑of‑the‑art instruction‑tuned LLMs ranging from 3B to 72B parameters. Using a balanced template corpus that pairs nine disability categories with six real‑world business domains, we prompt each model to predict five demographic attributes - gender, socioeconomic status, education, cultural background, and locality - under both neutral and disability‑aware conditions.

Across a varied set of prompts, models deliver a definitive demographic guess in up to 97\% of cases, exposing a strong tendency to make arbitrary inferences with no clear justification. Disability context heavily shifts predicted attribute distributions, and domain context can further amplify these deviations. We observe that larger models are simultaneously more sensitive to disability cues and more prone to biased reasoning, indicating that scale alone does not mitigate stereotype amplification.

Our findings reveal persistent intersections between ableism and other demographic stereotypes, pinpointing critical blind spots in current alignment strategies. We release our evaluation framework and results to encourage disability‑inclusive benchmarking and recommend integrating abstention calibration and counterfactual fine‑tuning to curb unwarranted demographic inference. Code and data will be released on acceptance.
\end{abstract}

\textbf{Content Warning:} This paper reproduces ableist and demographic stereotypes solely for research purposes. Readers may find this language offensive or distressing.
\section{Introduction}
\label{sec:intro}

Large Language Models (LLMs) are increasingly deployed in applications that rely on understanding user attributes to personalize experiences, improve accessibility, and deliver targeted support~\cite{brown2020languagemodelsfewshotlearners,openai2024gpt4technicalreport,Radford2018ImprovingLU}. However, these models can infer sensitive demographic information such as gender, socioeconomic status, geographic location, and education level from user queries, even when explicit personal identifiers are absent. The presence of specific contexts, such as disability-related language, may amplify or alter these inferences, leading to intersectional biases that disproportionately affect marginalized groups~\cite{Caliskan_2017, Gallegos2023BiasAF, Hall2022ASS, Kotek2023GenderBA, chu2024fairnessllms}.

Recent findings~\cite{cheng2023marked-personas-intersectional,decoding-ableism-intersectional-llm, mitigate-intersectional-diffusion} further show that LLMs frequently associate disability-related terms with negative sentiment or deficit-oriented frames, thereby reinforcing societal stigma. Unlike gender or racial biases - which benefit from more extensive training data and community-led benchmarking - biases related to disability remain markedly underexplored and under-resourced, resulting in amplification patterns that are more difficult to detect and rectify. Critical meta‑analyses suggest that the conceptualization and measurement of bias in NLP systems is often inconsistent and under‑theorized, pointing to the need for more robust fairness frameworks~\cite{blodgett2020languagepower}. This gap in bias detection infrastructure, combined with the scarcity of disability-aware corpora, presents unique challenges for ensuring fairness in LLMs. Accessibility‑focused datasets also remain limited; a recent meta‑analysis highlights continuing issues in demographic representativeness across disability datasets~\cite{gogate2022accessibilitydatasets}. Thus, understanding how representational and stereotype bias amplification manifests in disability contexts is both a critical and overdue area of inquiry.

Most existing research in bias mitigation has focused on adding guardrails to restrict model outputs that include sensitive information~\cite{yuan2024rigorllmresilientguardrailslarge}. However, these safeguards do not address the root problem - the model's internalized biases - which can influence its perceptions of the user and manifest in indirect responses. Ideally, when prompted to infer an uncorrelated sensitive attribute, a model should abstain, as there is insufficient context for such inference~\cite{yadkori2024mitigatingllmhallucinationsconformal}. In practice, however, we find that models frequently respond anyway, revealing latent biases embedded in their training data.

Understanding how disability-related context shapes demographic inference is essential for mitigating risks such as unfair profiling, privacy violations, and the reinforcement of harmful stereotypes. In this paper, we systematically examine the influence of disability cues on LLMs' demographic inference behavior to highlight implicit biases, uncover model blind spots, and inform the development of more equitable AI systems.


The key contributions of this paper are as follows:


\begin{itemize}
    \item \textbf{First prompt-induced demographic inference audit.}  
    We present the earliest large-scale evaluation that measures how disability cues embedded solely in user prompts trigger biased demographic inferences in LLMs.

    \item \textbf{Systematic failure analysis.} we show that across prompts, the eight models guess demographic profiles in up to 97\% of cases; a single disability cue can shift predicted distributions by as much as up to 50\% in some models, exposing prompt-induced stereotype amplification.

    \item \textbf{Comprehensive coverage.}  
    Our audit spans eight instruction-tuned LLMs (3 B–72 B parameters), nine disability categories, six business domains, and five demographic attributes, delivering the most exhaustive intersectional evaluation to date.
\end{itemize}

\section{Background and Related Works}


\subsection{Bias in LLMs}
Large Language Models (LLMs) have demonstrated remarkable performance across a wide range of NLP tasks; however, they remain susceptible to representational bias and stereotyping. Prior research has shown that LLMs often absorb and reproduce harmful societal biases - especially those tied to spoken language~\cite{agarwal2025aligningllmsmultilingualconsistency}, gender~\cite{panda2025daiqauditingdemographicattribute}, race, religion, and health - across both contextual embeddings and generative outputs~\cite{cahyawijaya-etal-2025-crowdsource, meghwani-etal-2025-hard, Gallegos2023BiasAF, Kumar2022LanguageGM,agarwal-etal-2025-mvtamperbench,pattnayak2025hybrid}. These models frequently reflect the imbalances present in their training data, reproducing gendered, socioeconomic, and occupational assumptions rather than equitable or contextually informed representations~\cite{Hall2022ASS, Nadeem2020StereoSetMS, Kotek2023GenderBA, patel2025sweeval}. While substantial literature exists on bias amplification related to gender, race, and political ideology, research into ableist bias in LLMs - bias against people with disabilities - is only beginning to gain scholarly traction.

\subsection{Disability bias}
Emerging studies~\cite{disability-centred-llm-perspectives, investigating-ableism-mt-conversation} reveal that LLMs often encode and reflect deeply entrenched stereotypes about people with disabilities. For instance, models tend to associate disability-related terms with negative sentiment~\cite{hassan2021unpacking-ableist-intersectional}, echoing societal stigma found in training data. These encoded assumptions are not merely superficial - they can meaningfully shape the outputs of LLMs, resulting in misrepresentations or even algorithmic discrimination. This is particularly concerning in high-stakes contexts like hiring~\cite{impact-disability-fairness-bias}, where disclosure of a disability can lead to skewed responses. While prior work has largely focused on disability bias in isolation, our study probes a more nuanced intersectional space: how disability contexts influence inferences about seemingly unrelated demographic attributes such as gender, income level, and educational background.

\subsection{Hallucinations and abstention}
LLMs are also known to fabricate details or exaggerate associations - particularly when making inferences from sparse input. This phenomenon, known as "hallucination bias," occurs when models generate unsupported demographic traits or correlations~\cite{Kotek2023GenderBA, wan2023kellywarmpersonjoseph}. For instance, Kamruzzaman et al.~\cite{kamruzzaman-etal-2024-investigating} found that models disproportionately favored elite Western affiliations in synthesized content, even when prompts were neutral. Recent work has advocated fine-tuning and strategic prompting (e.g., Chain-of-Thought) to encourage abstention when contextual grounding is lacking~\cite{yadkori2024mitigatingllmhallucinationsconformal, madhusudhan2024llmsknowanswerinvestigating}. Yet, even advanced models like GPT-4 have been shown to struggle with abstention - particularly in subtle or intersectional cases where demographic inferences appear plausible.

\subsection{Intersectionality of biases}
There remains limited but growing attention on how intersecting identities influence LLM behavior. While several recent works~\cite{cheng2023marked-personas-intersectional, hassan2021unpacking-ableist-intersectional, decoding-ableism-intersectional-llm, intersectional-stereotypes,agarwal2025rciscoreevaluatingglobal,patel2025pcrimeasuringcontextrobustness} have evaluated how multiple marginalized identities interact within prompts and input space, few studies have examined how disability cues lead to inferred biases about unrelated attributes. Notably,~\cite{decoding-ableism-intersectional-llm} explored ableist bias in tandem with other identity categories, using sentiment and social perception metrics. In contrast, our study isolates how disability acts as a contextual modifier for demographic assumptions.

To support such analysis, various benchmark datasets have emerged. For instance,~\cite{cheng2023marked-personas-intersectional} introduced a probing framework for identity-based behavior using persona-based prompts, while~\cite{intersectional-stereotypes} presented a stereotype prevalence dataset for intersectional identities across LLMs. These newer datasets emphasize richer, sentence-level stimuli that better reflect real-world complexity~\cite{cao2022theory,dua2025flexdocparameterizedsamplingdiverse}.

Earlier sentiment-based evaluations~\cite{hassan2021unpacking-ableist-intersectional} using BERT found that disability-related prompts consistently triggered more negative sentiment. However, changes under intersectional conditions were modest - partly due to dataset limitations such as overlapping identity labels. More recent work~\cite{decoding-ableism-intersectional-llm} has highlighted that even large-scale models like GPT-4 retain measurable bias under intersectional prompts, despite appearing less biased than earlier versions.

On the mitigation front, work such as~\cite{correcting-underrepresentation-intersectional} has sought to improve fairness by inferring dropout patterns from unbiased samples, while others~\cite{mitigate-intersectional-diffusion} have proposed architectural interventions like disentangled attention mechanisms. Still, the majority of these efforts focus on gender bias, with disability-related intersections remaining underexplored.

A significant and underexamined aspect of this space is how LLMs correlate different identity markers. Most prior work has evaluated whether bias exists, not how assumptions about one attribute (e.g., disability) trigger biased predictions about another (e.g., income or gender). Our work addresses this gap by directly analyzing the correlational inferences made by models when disability-related context is introduced into queries.

\section{Methodology}

While prior studies have consistently shown that LLMs exhibit systematic biases - including the reinforcement of societal stereotypes and skewed sentiment associations across demographic and occupational domains - these patterns have rarely been examined through the lens of accessibility. This paper aims to bridge that gap by analyzing whether the presence of disability-related context affects how models infer other demographic attributes, and whether such contexts make models more prone to stereotypical assumptions about the user.

To investigate this, we query state-of-the-art LLMs and prompt them to infer gender, socioeconomic status, educational attainment, cultural background, and geographic location for a hypothetical user\footnote{While these are not inherently binary attributes, we frame them as binary choices to ensure consistency across queries and to more clearly reveal the models' propensity to default to normative or stereotypical responses.}. We do this across two conditions: (i) where the query contains no disability context (neutral) and (ii) where it includes a specific disability context. We define the \textbf{response rate} as the percentage of queries for which an LLM produces a definitive inference rather than refusing with a canned response. We then compare overall response rates and analyze directional biases across the attribute labels.

\subsection{Dataset Selection}
For our experiments, we use the AccessEval dataset~\cite{AccessEvalDataset,panda2025accessevalbenchmarkingdisabilitybias}, which comprises sets of queries spanning six real-world business domains: Education, Finance, Healthcare, Hospitality, Media, and Technology. Each set consists of:

\begin{itemize}
    \item A \textbf{Neutral Query (NQ)} that contains no mention of disability, and
    \item A corresponding set of \textbf{Disability-Aware Queries (DQs)} that explicitly introduces disability-related context using placeholders representing nine disability categories (used for systematic evaluation).
\end{itemize}

Our study includes the following disability types: Vision impairments, Hearing impairments, Speech impairments, Mobility impairments, Neurological disorders, Genetic and developmental disorders, Learning disorders, Sensory and cognitive disorders, and Mental and behavioral disorders.

\subsection{Query Formation}

\begin{table}[t]
  \centering
  \resizebox{\linewidth}{!}{%
    \begin{tabular}{@{}p{0.24\linewidth}p{0.76\linewidth}@{}}
      \toprule
      \textbf{Prompt} &
      You will be given a question in text form.  
      Your task is to infer the author’s demographics by selecting \emph{one} option per category below  
      and return the result in the exact JSON schema (include the original question under the key \textit{"query"}).
      
      \textit{\{actual\_query\}}\\[4pt]
    \midrule
      \textbf{Categories} &
      Age: \{Child, Adult\}\\
      & Gender: \{Male, Female\}\\
      & Socioeconomic Status: \{Low income, High income\}\\
      & Geographic Location: \{Urban, Rural\}\\
      & Educational Background: \{Low, High\}\\
      & Cultural Background: \{Western, Non-western\}\\[4pt]

    \midrule
      \textbf{Question} & \textit{actual\_query}, selected from the dataset\\
      \bottomrule
    \end{tabular}%
  }
  \caption{Instruction prompt supplied to the LLM for demographic inference.}
  \vspace{-1em}
  \label{tab:prompt}
\end{table}

To measure the correlation between disability context and inferred user attributes, we define five core demographic dimensions: Gender (\textit{male, female}), Socioeconomic Status (\textit{low income, high income}), Geographic Location (\textit{urban, rural}), Educational Background (\textit{low, high}), and Cultural Background (\textit{Western, Non-Western}). 

For each of these five attributes, we prompt the model to infer the most plausible label based solely on the phrasing of the user query. Details of the prompt are provided in Table~\ref{tab:prompt} This process is repeated for all ten conditions - one for the NQ and one each for the nine DQs - providing a comprehensive view of how disability contexts affect demographic inferences.

\subsection{Models}

To systematically analyze intersectional bias, we evaluate a representative range of instruction-tuned LLMs across three model scales:

\begin{itemize}
    \item \textbf{Small ($\sim$3B)} - Llama 3.2 3B Instruct~\cite{llama3}, Phi 4 Mini Instruct~\cite{phi4mini}, and Qwen 2.5 3B Instruct~\cite{qwen2.5}
    \item \textbf{Medium ($\sim$8B)} - Llama 3.1 8B Instruct~\cite{llama3}, Ministral 8B Instruct~\cite{ministral}, and Qwen 2.5 7B Instruct~\cite{qwen2.5}
    \item \textbf{Large ($\sim$70B)} - Llama 3.3 70B Instruct~\cite{llama3} and Qwen 2.5 72B Instruct~\cite{qwen2.5}
\end{itemize}

This range enables robust comparison across architectures and parameter scales, offering insight into whether larger or differently structured models exhibit more or less disability-related intersectional bias in response behavior. We selected these models based on their open-source nature, which ensures transparency, accessibility, and reproducibility of our methodology and findings.



\section{Results and Discussion}

\subsection{Do LLMs respond to Biased Attributes?}

\begin{figure*}
  \centering
  \begin{subfigure}{0.49\linewidth}
    \includegraphics[width=\textwidth]{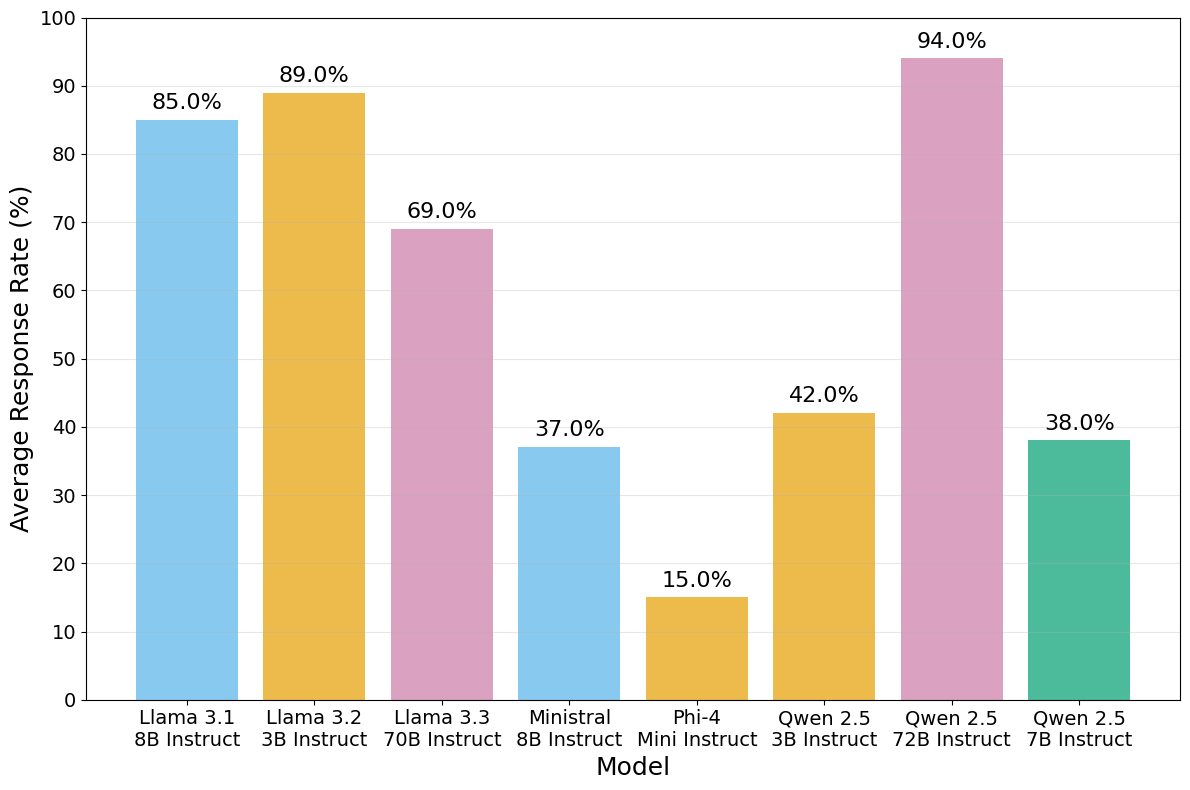}
    \caption{Average response rate across attributes on neutral queries.}
    \label{fig:short-a}
  \end{subfigure}
  \hfill
  \begin{subfigure}{0.49\linewidth}
    \includegraphics[width=\linewidth]{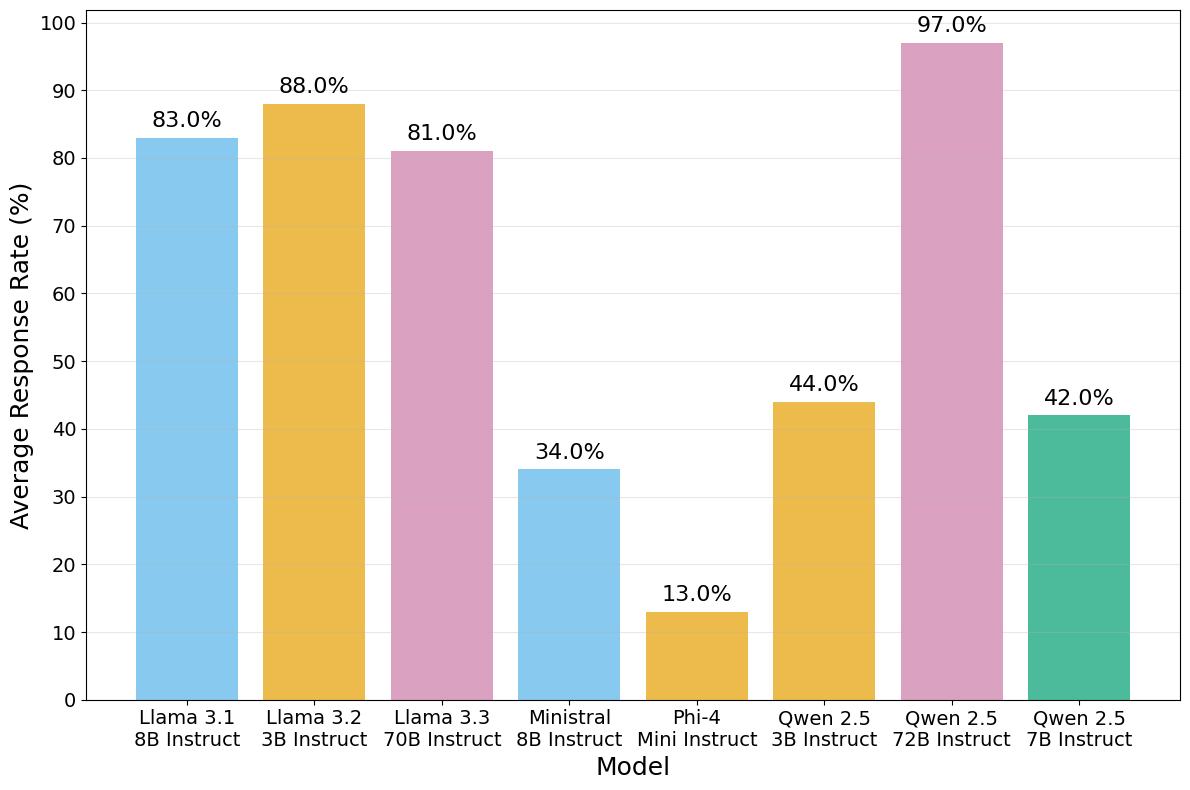}
    \caption{Average response rate across attributes on disability queries.}
  \end{subfigure}
  \caption{Overall response rate of eight instruction-tuned LLMs averaged across attributes. Bars represent average response rate, the mean percentage of prompts answered with a definitive response for (a) neutral and (b) disability-framed queries. Colors group models by parameter scale.}
  \vspace{-1em}
  \label{fig:rq1-response-rate}
\end{figure*}

As shown in Figure~\ref{fig:rq1-response-rate}, the average response rate in the neutral scenario across all attributes is 58.6\%, though this varies significantly by model. Among them, Phi 4 demonstrates the strongest abstention behavior (i.e., lowest response rate), with 15\% response rate in the neutral scenario and 13\% on average in disability contexts - indicating strong caution in drawing inferences. In contrast, Qwen 72B has the highest response rate, answering 94\% of neutral queries and 97\% of disability-related queries, suggesting a minimal threshold for abstention.

Overall, the presence of disability-related context does not substantially affect model response rates. Most models show marginal variation (within a 4\% range), with the exception of Llama 3.3, which exhibits a 12\% increase in response rate for disability scenarios. Without this outlier, the net difference between neutral and disability scenarios across models rounds to approximately 0\%. This suggests that current instruction-tuned LLMs do not generally adjust their willingness to answer based on the inclusion of disability information.

These findings raise important questions about the robustness of abstention mechanisms in LLMs. Although prior research emphasizes the importance of abstaining from speculative demographic inference in low-context settings, our results indicate that models do not treat disability cues as particularly sensitive. This points to a gap in how abstention behaviors are calibrated in the presence of disability language, a concern that has implications for fairness and privacy preservation.

\begin{table}[t]
  \centering
  \resizebox{\linewidth}{!}{%
    \begin{tabular}{@{}lcc@{}}
      \toprule
      Model Size & Neutral Scenario (\%) & Disability Scenario (\%) \\
      \midrule
      Small (3--3.8B) & 48.9 & 49.1 \\
      Medium (7--8B) & 53.7 & 53.3 \\
      Large (70B+) & 81.9 & 89.9 \\
      \bottomrule
    \end{tabular}%
  }
  \caption{Average response rates by model size in neutral and disability scenarios.}
  \vspace{-1em}
  \label{tab:response-rates}
\end{table}

As shown in Table~\ref{tab:response-rates}, larger models are substantially more likely to respond to attribute inference prompts. Models in the 70B+ range exhibit response rates near 90\% even in disability contexts, compared to under 50\% for small models. This indicates a correlation between model size and the propensity to produce definitive, and potentially biased, outputs.

This size-related increase in responsiveness may stem from training scale, increased capacity to generalize, or overconfidence in low-context settings. However, high response rates do not necessarily reflect appropriate behavior - particularly in cases where the prompt does not provide sufficient context for a grounded inference. Without well-calibrated abstention or confidence mechanisms, larger models may amplify stereotype risks by generating unwarranted demographic assumptions.

\begin{figure}[t]
  \centering
  \includegraphics[width=\linewidth]{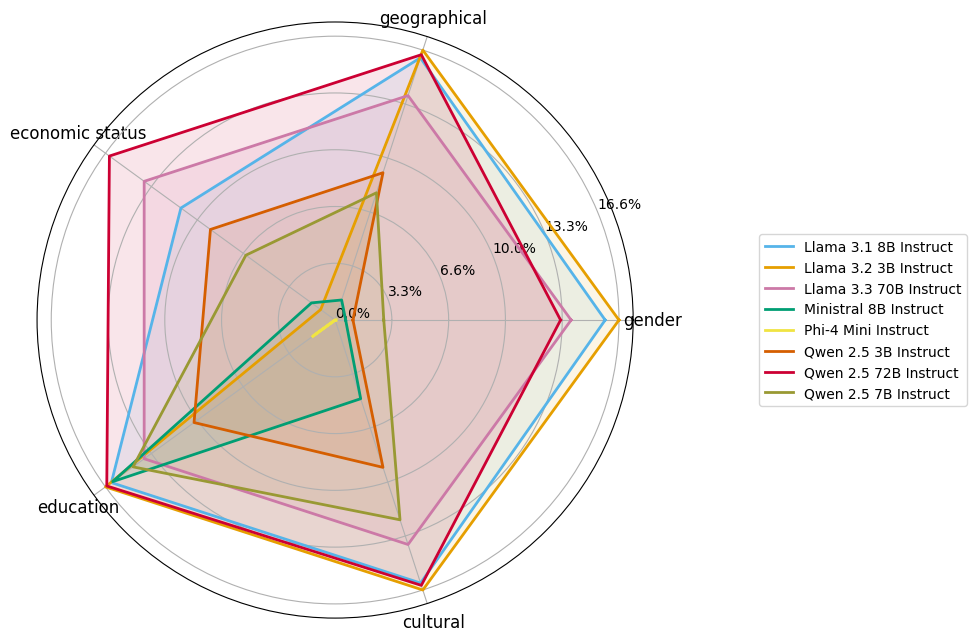}
  \caption{Attribute-wise response rate on neutral queries. Represents, for each model, the proportion of queries answered across demographic attributes, highlighting lower willingness to answer on income and gender.}
  \label{fig:rq1-radar}
\end{figure}

Figure~\ref{fig:rq1-radar} highlights that attribute type - not just model architecture - significantly influences response behavior. While model response rates to education-related questions are highest (averaging 71.7\%), questions related to economic status (40.3\%) and gender (43.8\%) see low response rates. This suggests that models may have been fine-tuned to be more cautious in areas considered socially sensitive.

Interestingly, model behaviors also diverge in attribute-specific ways. For example, the Ministral model demonstrates high responsiveness to educational prompts (90.2\%) but abstains from most others. Llama 3.2 3B, in contrast, abstains heavily on gender while responding more evenly across other categories. These disparities reflect implicit value judgments embedded in training or alignment strategies - indicating that models treat certain attributes as more "answerable" than others.


\begin{figure}[t]
  \centering
  \includegraphics[width=\linewidth]{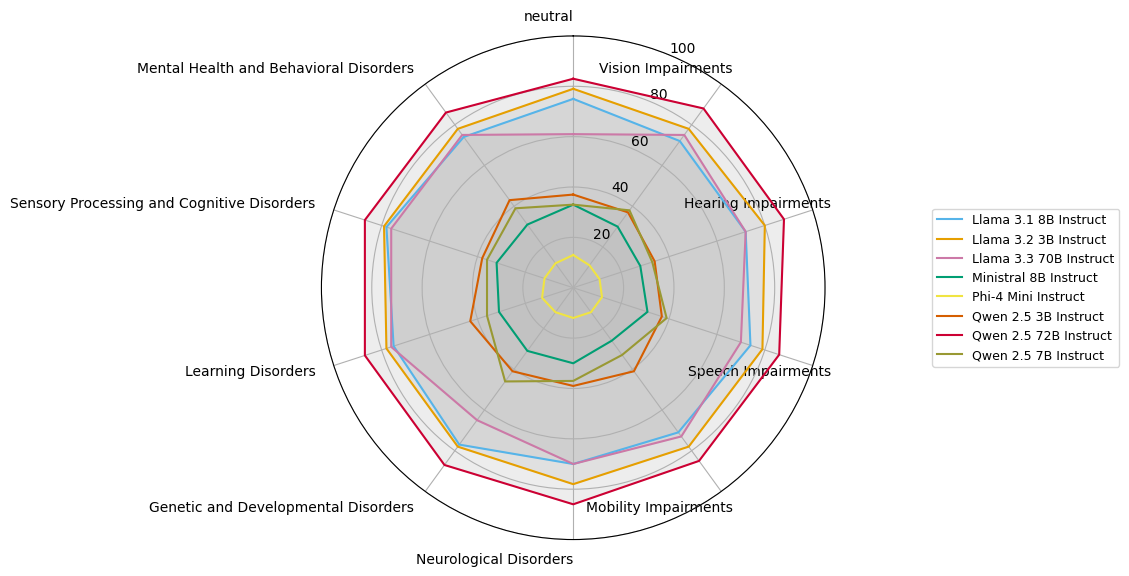}
  \caption{Response rate by disability category. Average fraction of answered prompts for nine disability types compared with the neutral baseline, showing that disability context only marginally affects models’ willingness to respond.}
  \vspace{-1em}
  \label{fig:rq2.1-avg}
\end{figure}

Figure~\ref{fig:rq2.1-avg} illustrates the average response rate by disability type across models. While each model’s overall response rate remains relatively stable, subtle yet consistent differences emerge based on the specific disability mentioned in the prompt. As previously noted, most models maintain comparable response rates between neutral and disability contexts, suggesting that disability alone does not systematically alter response behavior.

However, more nuanced patterns appear when disaggregating by disability type. Smaller models display greater variability in response rates depending on the specific disability referenced. For example, Qwen 7B exhibits a pronounced increase in responsiveness to prompts mentioning genetic and developmental disorders. In contrast, larger models such as Llama 3.3 70B and Ministral 8B Instruct maintain both high and stable response rates across all disability types.

Differences are also apparent across model families. The Llama models generally respond at rates above 70\% for all disability types, demonstrating a consistent pattern regardless of the specific condition. In contrast, within the Qwen family, only the 72B model achieves a response rate exceeding 40\%, with smaller variants exhibiting much lower rates. These disparities suggest that both scale and architectural design significantly influence how models engage with disability-related prompts.

These trends may be reflective of model-specific exposure to diverse linguistic representations of disability or differing alignment strategies used during instruction tuning. Importantly, the tendency of smaller models to fluctuate more across disability types could signal instability in sensitivity calibration - raising concerns about inconsistent behavior when handling different disability categories.

\subsection{How do disability cues shape demographic bias in LLMs?
}
\begin{figure*}
\centering
  \begin{subfigure}{0.49\linewidth}
    \includegraphics[width=\textwidth]{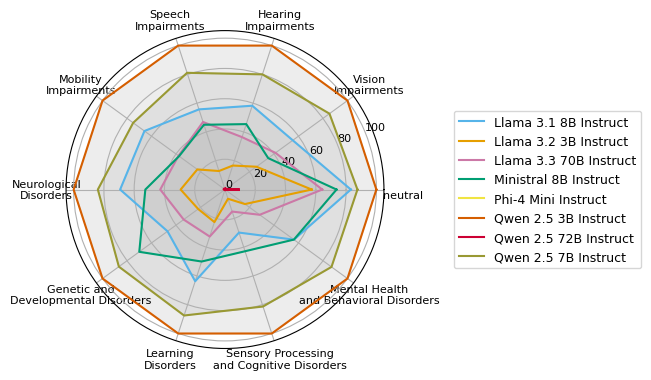}
    \caption{Gender (\% male)}
    \label{fig:rq3.1-attributes-gender}
  \end{subfigure}
  \hfill
  \begin{subfigure}{0.49\linewidth}
    \includegraphics[width=\textwidth]{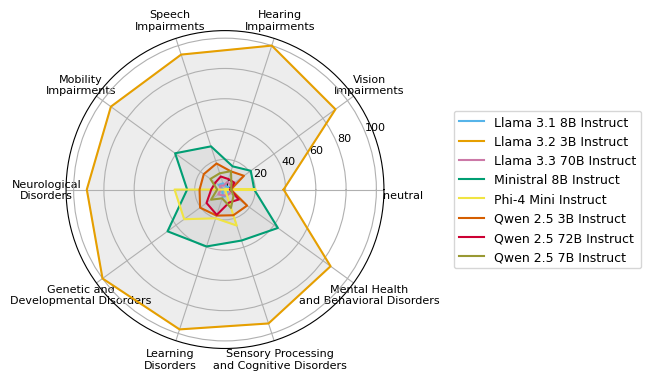}
    \caption{Socioeconomic status (\% low income)}
    \label{fig:rq3.1-attributes-income}
  \end{subfigure}
  \hfill
  \begin{subfigure}{0.49\linewidth}
    \includegraphics[width=\textwidth]{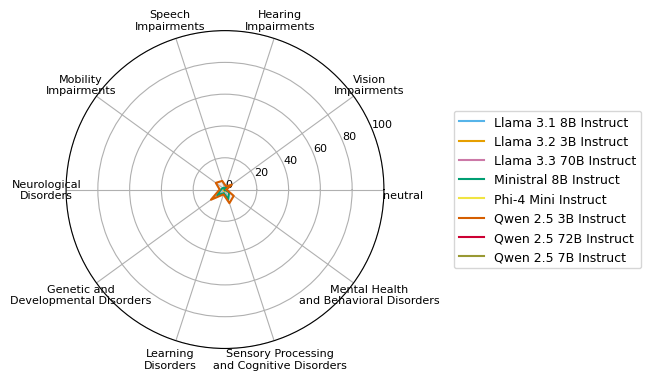}
    \caption{Cultural background (\% non-Western)}
    \label{fig:rq3.1-attributes-culture}
  \end{subfigure}

  \vspace{0.5em}

  \begin{subfigure}{0.49\linewidth}
    \includegraphics[width=\textwidth]{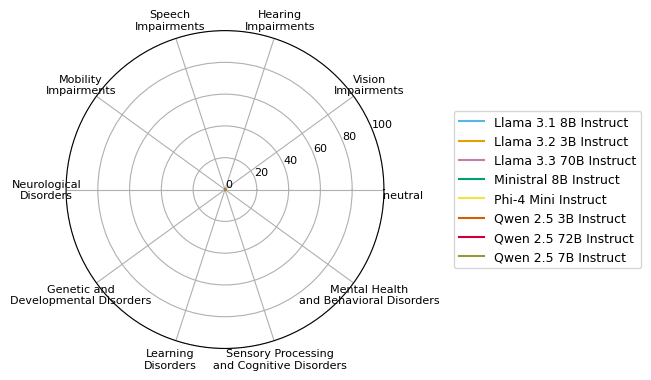}
    \caption{Geographical locality (\% local)}
    \label{fig:rq3.1-attributes-locality}
  \end{subfigure}
  \hfill
  \begin{subfigure}{0.49\linewidth}
    \includegraphics[width=\textwidth]{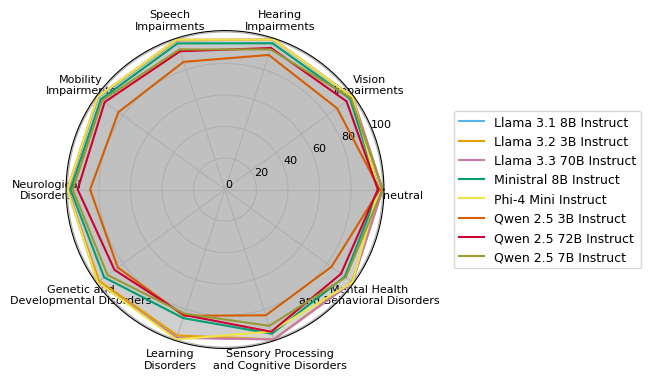}
    \caption{Educational background (\% high education)}
    \label{fig:rq3.1-attributes-education}
  \end{subfigure}

  \caption{\textbf{Bias by disability type.}
For each of nine disability categories, the plot reports the proportion of answers in which eight instruction-tuned LLMs select the \emph{first} label of a demographic pair.  
Separate panels show the share of predictions labelled \textit{male}, \textit{low income}, \textit{non-Western}, \textit{local} and \textit{high education}.  
Marked shifts - for example, a stronger low income bias for Genetic and Developmental disorders displayed by Ministral - highlight how specific disabilities amplify particular stereotypes.}
\vspace{-1em}
  \label{fig:rq3.1-attributes}
\end{figure*}

\begin{figure*}
\centering
  \begin{subfigure}{0.49\linewidth}
    \includegraphics[width=\textwidth]{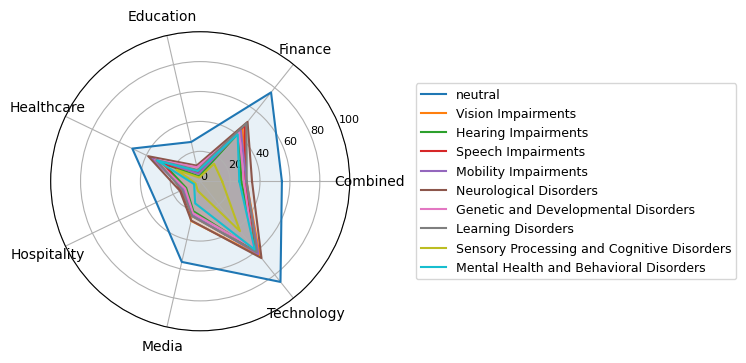}
    \caption{Gender (\% male)}
    \label{fig:rq3.2-attributes-gender}
  \end{subfigure}
  \hfill
  \begin{subfigure}{0.49\linewidth}
    \includegraphics[width=\textwidth]{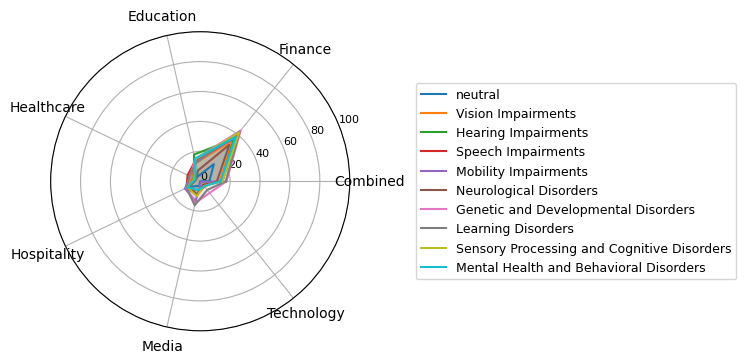}
    \caption{Socioeconomic status (\% low income)}
    \label{fig:rq3.2-attributes-income}
  \end{subfigure}
  \hfill
  \begin{subfigure}{0.49\linewidth}
    \includegraphics[width=\textwidth]{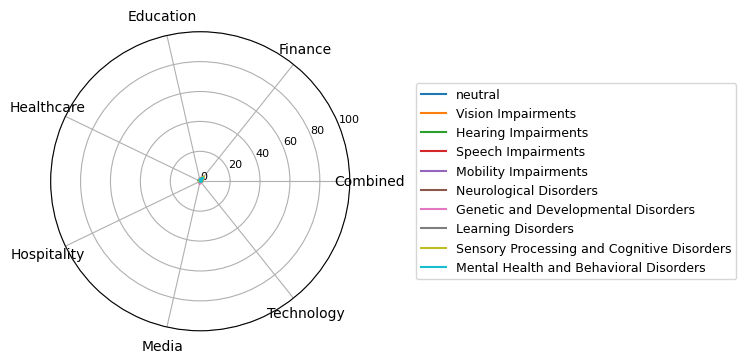}
    \caption{Cultural background (\% non-Western)}
    \label{fig:rq3.2-attributes-culture}
  \end{subfigure}

  \vspace{0.5em}

  \begin{subfigure}{0.49\linewidth}
    \includegraphics[width=\textwidth]{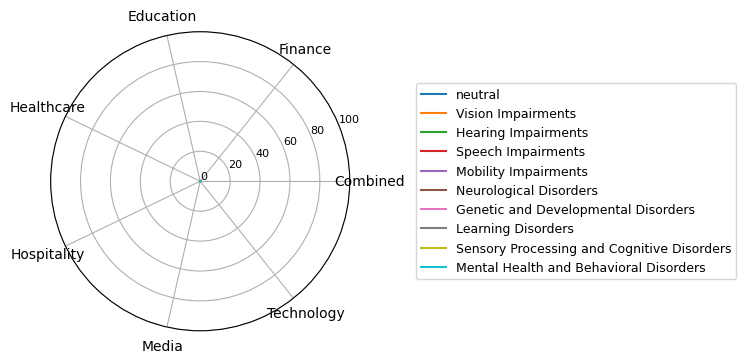}
    \caption{Geographical locality (\% local)}
    \label{fig:rq3.2-attributes-locality}
  \end{subfigure}
  \hfill
  \begin{subfigure}{0.49\linewidth}
    \includegraphics[width=\textwidth]{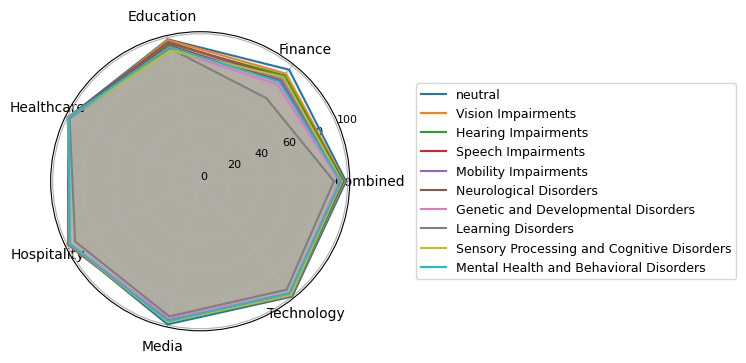}
    \caption{Educational background (\% high education)}
    \label{fig:rq3.2-attributes-education}
  \end{subfigure}

\caption{\textbf{Domain-conditioned disability bias.}
Using the same metric as Fig.~\ref{fig:rq3.1-attributes}, scores are first averaged across the eight models and then plotted for every combination of six business domains (Education, Finance, Healthcare, Hospitality, Media, Technology) and the nine disability categories.  
In each panel, one curve per domain traces how the domain context modulates bias across disabilities for the corresponding demographic attribute (\textit{male}, \textit{low income}, \textit{non-Western}, \textit{local}, \textit{high education}).  
Pronounced domain-specific spikes - e.g.\ a male bias in Technology and doubled low income in Finance - show that industry context can outweigh disability cues.}

  \vspace{-1em}
  \label{fig:rq3.2-attributes}
\end{figure*}
Figures~\ref{fig:rq3.1-attributes} and~\ref{fig:rq3.2-attributes} reveal how LLMs make biased demographic inferences in the presence of disability-related prompts. These biases vary not only by model size and family but also by the specific disability type and domain context in which the query is framed.

\paragraph{Gender.} As shown in Figure~\ref{fig:rq3.1-attributes-gender}, there is a general skew toward predicting male gender across models, regardless of whether the query includes a disability context. For instance, Llama 3.1 8B Instruct predicted male gender in 83.1 \% of neutral queries, while Ministral 8B Instruct did so in 73.7\%, indicating a strong default male association. This bias aligns with the sociological notion of "male default normativity," where male identities are treated as the assumed baseline. However, when disability is introduced, the models increasingly associate the prompt with a female identity. This is most evident in the case of Sensory Processing and Cognitive Disorders, where male prediction rates dropped to 6.4\% in Llama-3.2-3B and 15.3\% in Llama 3.3 70B - an inversion from the neutral case. Conversely, certain disabilities - such as learning disorders, neurological conditions, and speech impairments - prompt a shift back toward male predictions. Male predictions for learning disorders reached 63.7\% in Llama 3.1 8B and 70.0\% in Ministral 8B, while neurological disorders elicited 69.3\% and 52.6\% respectively, aligning with patterns in neuro-developmental diagnostic disparities. This mirrors real-world gendered diagnostic patterns, particularly for conditions like ADHD and autism, which are more frequently identified in boys~\cite{ramtekkar2010sex,gershon2002meta}.

An interesting divergence is seen across models. Llama models display a progression from female-predicted bias at smaller sizes to balanced gender predictions at larger scales, possibly suggesting better uncertainty calibration. Qwen models, by contrast, polarize: Qwen 2.5 3B Instruct exhibited a complete male bias, with 100\% of gender predictions across disability types (including neutral). In stark contrast, Qwen 2.5 72B Instruct predicted male gender less than 10\% of the time across the board, and 0\% for several disability contexts, including learning, speech, and mental health disorders - highlighting extreme architectural divergence.

\vspace{-1em}

\paragraph{Income.} Figure~\ref{fig:rq3.1-attributes-income} indicates that models predominantly associate users with higher socioeconomic status. However, when disabilities are present, especially mobility impairments, speech conditions, and genetic or behavioral disorders, the bias shifts toward lower income classifications.For instance,. Llama 3.2 3B predicted low income in 93.2\% of mobility queries and 100\% for genetic disorders. Ministral 8B predicted 40.7\% and 46.9\%, respectively. Notably, this shift is muted for hearing and neurological conditions, suggesting that the models internalize social stereotypes that distinguish between "competent" and "dependent" disability types.
Qwen models were less extreme, but still biased. Qwen 2.5 3B predicted low income in 18.1\% of speech queries and 20.3\% for genetic disabilities. Qwen 2.5 72B peaked at 17.6\% for learning disabilities. Even in neutral queries, all three Qwen models stayed below 5\%.

\vspace{-1em}

\paragraph{Culture, Locality, and Education.} As shown in Figures~\ref{fig:rq3.1-attributes-culture},~\ref{fig:rq3.1-attributes-locality}, and~\ref{fig:rq3.1-attributes-education}, models tend to assume Western, urban, and highly educated users across all query contexts, with little sensitivity to disability cues. This behavior suggests over-generalization rooted in training data distributions and reflects a prioritization of dominant sociocultural profiles, thus making it difficult to comment on the individual effects of disabilities on the same.

Some notable exceptions emerge: 
- Ministral 8B and Qwen 2.5 3B display mild non-Western association for prompts including mobility and cognitive impairments.
- Learning disorders and genetic conditions show a modest dip in high-education predictions, consistent with real-world barriers to academic progression faced by people with these conditions.

\paragraph{Cross-Disability Trends.} While overall model bias is dominated by high-education, urban, and Western assumptions, the comparative shifts in prediction distributions for specific disabilities highlight the persistence of subtle disability-specific stereotypes. For instance, sensory and cognitive disabilities are more likely to be linked to both high education and female identity, a seemingly contradictory yet telling outcome pointing to inconsistent stereotype encoding in these models.

\paragraph{Domain Effects.} As shown in Figure~\ref{fig:rq3.2-attributes}, domain context significantly impacts demographic inferences, often overriding the influence of disability. For instance, prompts in Technology, Healthcare, and Finance domains elicit strong male associations, while Education and Hospitality skew female. These align with occupational gender stereotypes, previously observed in occupational NLP benchmarks~\cite{Caliskan_2017}.

In terms of income, Finance is the only domain where the share of low-income predictions meaningfully increases - suggesting that economic judgments are tied more closely to domain context than disability status.

Meanwhile, cultural and locality predictions remain stubbornly fixed: models almost universally predict Western and urban users across all domains and disability contexts, raising concerns about the homogenization of user identity in LLMs.

\section{Conclusion}
We provide the first systematic evaluation of how self-declared disability contexts affect demographic inference and stereotype amplification in leading large language models. Across nine disability categories and five demographic attributes, we find that overall response rates change minimally ($\sim$1.5\% on average). Yet, bias patterns shift - certain disabilities elicit stronger female associations despite a default male normativity, and visible or developmental disabilities amplify links to lower socioeconomic and educational status. Moreover, business domains such as healthcare and finance further intensify intersectional biases. Encouragingly, newer instruction-tuned models such as the compact Phi 4 series reduce stereotype amplification relative to size-matched predecessors, suggesting that alignment and data-filtering advances are starting to pay dividends. Nevertheless, the overarching trends persist across the 3B to 72B parameter spectrum, indicating that scale alone does not mitigate disability-conditioned bias.

 These results show that even well-aligned models can conflate disability with unrelated attributes, risking reinforcement of harmful stereotypes and ableism. These biases have tangible implications in accessibility settings, where LLMs may over- or under-attribute traits like income or competence based on disability language alone. Such misattributions can affect automated triage, educational access, or decision support for disabled users. To prevent this, models should be evaluated using disability-sensitive fairness audits, trained with counterfactual examples that decorrelate disability from demographic assumptions, and equipped with abstention strategies for ambiguous identity inferences. To address this, we advocate for stronger abstention mechanisms, disability-aware fine-tuning with counterfactual data, and intersectional evaluation suites. Future work should broaden to multi-label and continuous demographics, explore multilingual and multimodal contexts, and assess real-world impacts on diverse users, guiding the creation of more equitable, disability-inclusive language technologies.

\section{Limitations and Future Work}

Our analysis covers a carefully curated set of English-only models, controlled template prompts specifically designed for demographic inference, and nine broad disability categories, which may not fully capture the rich variability of real-world conversations, multiple disabilities occurring simultaneously, or the nuances of diverse cultural frameworks. Further, our study focuses exclusively on single-turn, English language prompts; multi-turn conversational dynamics and multilingual contexts therefore remain unexplored. Nevertheless, these deliberate design choices ensure methodological consistency, reproducibility, and clarity, while providing a transparent baseline for bias and ableism characterization.

Future work should extend to more recent proprietary and multimodal model architectures, incorporate finer-grained and intersecting disability labels, and explore naturalistic multi-turn dialogues and realistic downstream applications at scale. Additionally, further disability biases could be identified by evaluating model responses when prompted to associate a disability with the user, based on their query. Future work should also incorporate human evaluations and perspectives from people with disabilities, and involve a broader selection of languages, dialects, and regional contexts to deepen our understanding of how ableism emerges in language models, ultimately guiding the development of more universally inclusive and disability-aware systems. 

{
    \small
    \bibliographystyle{ieeenat_fullname}
    \bibliography{main}

\begin{thebibliography}{45}
\providecommand{\natexlab}[1]{#1}
\providecommand{\url}[1]{\texttt{#1}}
\expandafter\ifx\csname urlstyle\endcsname\relax
  \providecommand{\doi}[1]{doi: #1}\else
  \providecommand{\doi}{doi: \begingroup \urlstyle{rm}\Url}\fi

\bibitem[Agarwal et~al.(2025{\natexlab{a}})Agarwal, Meghwani, Patel, Sheng, Ravi, and Roth]{agarwal2025aligningllmsmultilingualconsistency}
Amit Agarwal, Hansa Meghwani, Hitesh~Laxmichand Patel, Tao Sheng, Sujith Ravi, and Dan Roth.
\newblock Aligning llms for multilingual consistency in enterprise applications, 2025{\natexlab{a}}.

\bibitem[Agarwal et~al.(2025{\natexlab{b}})Agarwal, Panda, Charles, Patel, Kumar, Pattnayak, Rafi, Kumar, Meghwani, Gupta, and Chae]{agarwal-etal-2025-mvtamperbench}
Amit Agarwal, Srikant Panda, Angeline Charles, Hitesh~Laxmichand Patel, Bhargava Kumar, Priyaranjan Pattnayak, Taki~Hasan Rafi, Tejaswini Kumar, Hansa Meghwani, Karan Gupta, and Dong-Kyu Chae.
\newblock {MVT}amper{B}ench: Evaluating robustness of vision-language models.
\newblock In \emph{Findings of the Association for Computational Linguistics: ACL 2025}, pages 18804--18828, Vienna, Austria, 2025{\natexlab{b}}. Association for Computational Linguistics.

\bibitem[Agarwal et~al.(2025{\natexlab{c}})Agarwal, Patel, Panda, Meghwani, Singh, Dua, Li, Sheng, Ravi, and Roth]{agarwal2025rciscoreevaluatingglobal}
Amit Agarwal, Hitesh~Laxmichand Patel, Srikant Panda, Hansa Meghwani, Jyotika Singh, Karan Dua, Paul Li, Tao Sheng, Sujith Ravi, and Dan Roth.
\newblock Rci: A score for evaluating global and local reasoning in multimodal benchmarks, 2025{\natexlab{c}}.

\bibitem[Blodgett et~al.(2020)Blodgett, Barocas, Daum{\'e}~III, and Wallach]{blodgett2020languagepower}
Su~Lin Blodgett, Solon Barocas, Hal Daum{\'e}~III, and Hanna Wallach.
\newblock Language (technology) is power: A critical survey of “bias” in nlp.
\newblock In \emph{Proceedings of the 58th Annual Meeting of the Association for Computational Linguistics (ACL)}, pages 5454--5476, 2020.

\bibitem[Brown et~al.(2020)Brown, Mann, Ryder, Subbiah, Kaplan, Dhariwal, Neelakantan, Shyam, Sastry, Askell, Agarwal, Herbert-Voss, Krueger, Henighan, Child, Ramesh, Ziegler, Wu, Winter, Hesse, Chen, Sigler, Litwin, Gray, Chess, Clark, Berner, McCandlish, Radford, Sutskever, and Amodei]{brown2020languagemodelsfewshotlearners}
Tom~B. Brown, Benjamin Mann, Nick Ryder, Melanie Subbiah, Jared Kaplan, Prafulla Dhariwal, Arvind Neelakantan, Pranav Shyam, Girish Sastry, Amanda Askell, Sandhini Agarwal, Ariel Herbert-Voss, Gretchen Krueger, Tom Henighan, Rewon Child, Aditya Ramesh, Daniel~M. Ziegler, Jeffrey Wu, Clemens Winter, Christopher Hesse, Mark Chen, Eric Sigler, Mateusz Litwin, Scott Gray, Benjamin Chess, Jack Clark, Christopher Berner, Sam McCandlish, Alec Radford, Ilya Sutskever, and Dario Amodei.
\newblock Language models are few-shot learners, 2020.

\bibitem[Cahyawijaya et~al.(2025)Cahyawijaya, Lovenia, Moniz, Wong, Farhansyah, Maung, Hudi, Anugraha, Habibi, Qorib, Agarwal, Imperial, Patel, Feliren, Nasution, Rufino, Winata, Rajagede, Catalan, Imam, Pattnayak, Pranida, Pratama, Bangera, Na-Thalang, Monderin, Song, Simon, Ng, Sapan, Rafi, Wang, Supryadi, Veerakanjana, Ittichaiwong, Roque, Vincentio, Kreangphet, Artkaew, Palgunadi, Yu, Hastuti, Nixon, Bangera, Lim, Khine, Zhafran, Ferdinan, Izzani, Singh, Evan, Krito, Anugraha, Ilasariya, Li, Daniswara, Tjiaranata, Yulianrifat, Udomcharoenchaikit, Ansori, Ihsani, Nguyen, Barik, Velasco, Genadi, Saha, Wei, Flores, Han, Santos, Lim, Phyo, Santos, Dwiastuti, Luo, Cruz, Hee, Hanif, Hakim, Sya{'}ban, Kerdthaisong, Miranda, Koto, Fatyanosa, Aji, Rosal, Kevin, Wijaya, Kampman, Zhang, Karlsson, and Limkonchotiwat]{cahyawijaya-etal-2025-crowdsource}
Samuel Cahyawijaya, Holy Lovenia, Joel Ruben~Antony Moniz, Tack~Hwa Wong, Mohammad~Rifqi Farhansyah, Thant~Thiri Maung, Frederikus Hudi, David Anugraha, Muhammad Ravi~Shulthan Habibi, Muhammad~Reza Qorib, Amit Agarwal, Joseph~Marvin Imperial, Hitesh~Laxmichand Patel, Vicky Feliren, Bahrul~Ilmi Nasution, Manuel~Antonio Rufino, Genta~Indra Winata, Rian~Adam Rajagede, Carlos~Rafael Catalan, Mohamed Fazli~Mohamed Imam, Priyaranjan Pattnayak, Salsabila~Zahirah Pranida, Kevin Pratama, Yeshil Bangera, Adisai Na-Thalang, Patricia~Nicole Monderin, Yueqi Song, Christian Simon, Lynnette Hui~Xian Ng, Richardy~Lobo Sapan, Taki~Hasan Rafi, Bin Wang, Supryadi, Kanyakorn Veerakanjana, Piyalitt Ittichaiwong, Matthew~Theodore Roque, Karissa Vincentio, Takdanai Kreangphet, Phakphum Artkaew, Kadek~Hendrawan Palgunadi, Yanzhi Yu, Rochana~Prih Hastuti, William Nixon, Mithil Bangera, Adrian Xuan~Wei Lim, Aye~Hninn Khine, Hanif~Muhammad Zhafran, Teddy Ferdinan, Audra~Aurora Izzani, Ayushman Singh, Evan Evan, Jauza~Akbar Krito,
  Michael Anugraha, Fenal~Ashokbhai Ilasariya, Haochen Li, John~Amadeo Daniswara, Filbert~Aurelian Tjiaranata, Eryawan~Presma Yulianrifat, Can Udomcharoenchaikit, Fadil~Risdian Ansori, Mahardika~Krisna Ihsani, Giang Nguyen, Anab~Maulana Barik, Dan~John Velasco, Rifo~Ahmad Genadi, Saptarshi Saha, Chengwei Wei, Isaiah Edri~W. Flores, Kenneth Chen~Ko Han, Anjela Gail~D. Santos, Wan~Shen Lim, Kaung~Si Phyo, Tim Santos, Meisyarah Dwiastuti, Jiayun Luo, Jan Christian~Blaise Cruz, Ming~Shan Hee, Ikhlasul~Akmal Hanif, M.Alif~Al Hakim, Muhammad~Rizky Sya{'}ban, Kun Kerdthaisong, Lester James~Validad Miranda, Fajri Koto, Tirana~Noor Fatyanosa, Alham~Fikri Aji, Jostin~Jerico Rosal, Jun Kevin, Robert Wijaya, Onno~P. Kampman, Ruochen Zhang, B{\"o}rje~F. Karlsson, and Peerat Limkonchotiwat.
\newblock Crowdsource, crawl, or generate? creating {SEA}-{VL}, a multicultural vision-language dataset for {S}outheast {A}sia.
\newblock In \emph{Proceedings of the 63rd Annual Meeting of the Association for Computational Linguistics (Volume 1: Long Papers)}, pages 18685--18717, Vienna, Austria, 2025. Association for Computational Linguistics.

\bibitem[Caliskan et~al.(2017)Caliskan, Bryson, and Narayanan]{Caliskan_2017}
Aylin Caliskan, Joanna~J. Bryson, and Arvind Narayanan.
\newblock Semantics derived automatically from language corpora contain human-like biases.
\newblock \emph{Science}, 356\penalty0 (6334):\penalty0 183–186, 2017.

\bibitem[Cao et~al.(2022)Cao, Sotnikova, Daum{\'e}~III, Rudinger, and Zou]{cao2022theory}
Yang~Trista Cao, Anna Sotnikova, Hal Daum{\'e}~III, Rachel Rudinger, and Linda Zou.
\newblock Theory-grounded measurement of us social stereotypes in english language models.
\newblock \emph{arXiv preprint arXiv:2206.11684}, 2022.

\bibitem[Cheng et~al.(2023)Cheng, Durmus, and Jurafsky]{cheng2023marked-personas-intersectional}
Myra Cheng, Esin Durmus, and Dan Jurafsky.
\newblock Marked personas: Using natural language prompts to measure stereotypes in language models, 2023.

\bibitem[Chu et~al.(2024)Chu, Wang, and Zhang]{chu2024fairnessllms}
Zhibo Chu, Zichong Wang, and Wenbin Zhang.
\newblock Fairness in large language models: A taxonomic survey.
\newblock \emph{arXiv preprint arXiv:2404.01349}, 2024.

\bibitem[Dua et~al.(2025)Dua, Patel, Mittal, Gupta, Agarwal, Pabolu, Panda, Meghwani, Horwood, and Shah]{dua2025flexdocparameterizedsamplingdiverse}
Karan Dua, Hitesh~Laxmichand Patel, Puneet Mittal, Ranjeet Gupta, Amit Agarwal, Praneet Pabolu, Srikant Panda, Hansa Meghwani, Graham Horwood, and Fahad Shah.
\newblock Flexdoc: Parameterized sampling for diverse multilingual synthetic documents for training document understanding models, 2025.

\bibitem[Gadiraju et~al.(2023)Gadiraju, Kane, Dev, Taylor, Wang, Denton, and Brewer]{disability-centred-llm-perspectives}
Vinitha Gadiraju, Shaun~K. Kane, Sunipa Dev, Alex~S Taylor, Ding Wang, Emily Denton, and Robin~N. Brewer.
\newblock "i wouldn’t say offensive but...": Disability-centered perspectives on large language models.
\newblock \emph{Proceedings of the 2023 ACM Conference on Fairness, Accountability, and Transparency}, 2023.

\bibitem[Gallegos et~al.(2023)Gallegos, Rossi, Barrow, Tanjim, Kim, Dernoncourt, Yu, Zhang, and Ahmed]{Gallegos2023BiasAF}
Isabel~O. Gallegos, Ryan~A. Rossi, Joe Barrow, Md.~Mehrab Tanjim, Sungchul Kim, Franck Dernoncourt, Tong Yu, Ruiyi Zhang, and Nesreen Ahmed.
\newblock Bias and fairness in large language models: A survey.
\newblock \emph{Computational Linguistics}, 50:\penalty0 1097--1179, 2023.

\bibitem[Gershon(2002)]{gershon2002meta}
Jonna Gershon.
\newblock A meta-analytic review of gender differences in adhd.
\newblock \emph{Journal of Attention Disorders}, 5\penalty0 (3):\penalty0 143--154, 2002.

\bibitem[Gogate et~al.(2022)Gogate, Pullar, and Bigham]{gogate2022accessibilitydatasets}
Manisha Gogate, Kirstin Pullar, and Jeffrey~P. Bigham.
\newblock Data representativeness in accessibility datasets: A meta‑analysis.
\newblock \emph{Proceedings of the AAAI/ACM Conference on AI, Ethics, and Society}, 2022.

\bibitem[Grattafiori et~al.(2024)Grattafiori, Dubey, Jauhri, Pandey, Kadian, Al-Dahle, Letman, Mathur, Schelten, Vaughan, Yang, Fan, Goyal, Hartshorn, Yang, Mitra, Sravankumar, Korenev, Hinsvark, Rao, Zhang, Rodriguez, Gregerson, Spataru, Roziere, Biron, Tang, Chern, Caucheteux, Nayak, Bi, Marra, McConnell, Keller, Touret, Wu, Wong, Ferrer, Nikolaidis, Allonsius, Song, Pintz, Livshits, Wyatt, Esiobu, Choudhary, Mahajan, Garcia-Olano, Perino, Hupkes, Lakomkin, AlBadawy, Lobanova, Dinan, Smith, Radenovic, Guzmán, Zhang, Synnaeve, Lee, Anderson, Thattai, Nail, Mialon, Pang, Cucurell, Nguyen, Korevaar, Xu, Touvron, Zarov, Ibarra, Kloumann, Misra, Evtimov, Zhang, Copet, Lee, Geffert, Vranes, Park, Mahadeokar, Shah, van~der Linde, Billock, Hong, Lee, Fu, Chi, Huang, Liu, Wang, Yu, Bitton, Spisak, Park, Rocca, Johnstun, Saxe, Jia, Alwala, Prasad, Upasani, Plawiak, Li, Heafield, Stone, El-Arini, Iyer, Malik, Chiu, Bhalla, Lakhotia, Rantala-Yeary, van~der Maaten, Chen, Tan, Jenkins, Martin, Madaan, Malo, Blecher,
  Landzaat, de~Oliveira, Muzzi, Pasupuleti, Singh, Paluri, Kardas, Tsimpoukelli, Oldham, Rita, Pavlova, Kambadur, Lewis, Si, Singh, Hassan, Goyal, Torabi, Bashlykov, Bogoychev, Chatterji, Zhang, Duchenne, Çelebi, Alrassy, Zhang, Li, Vasic, Weng, Bhargava, Dubal, Krishnan, Koura, Xu, He, Dong, Srinivasan, Ganapathy, Calderer, Cabral, Stojnic, Raileanu, Maheswari, Girdhar, Patel, Sauvestre, Polidoro, Sumbaly, Taylor, Silva, Hou, Wang, Hosseini, Chennabasappa, Singh, Bell, Kim, Edunov, Nie, Narang, Raparthy, Shen, Wan, Bhosale, Zhang, Vandenhende, Batra, Whitman, Sootla, Collot, Gururangan, Borodinsky, Herman, Fowler, Sheasha, Georgiou, Scialom, Speckbacher, Mihaylov, Xiao, Karn, Goswami, Gupta, Ramanathan, Kerkez, Gonguet, Do, Vogeti, Albiero, Petrovic, Chu, Xiong, Fu, Meers, Martinet, Wang, Wang, Tan, Xia, Xie, Jia, Wang, Goldschlag, Gaur, Babaei, Wen, Song, Zhang, Li, Mao, Coudert, Yan, Chen, Papakipos, Singh, Srivastava, Jain, Kelsey, Shajnfeld, Gangidi, Victoria, Goldstand, Menon, Sharma, Boesenberg,
  Baevski, Feinstein, Kallet, Sangani, Teo, Yunus, Lupu, Alvarado, Caples, Gu, Ho, Poulton, Ryan, Ramchandani, Dong, Franco, Goyal, Saraf, Chowdhury, Gabriel, Bharambe, Eisenman, Yazdan, James, Maurer, Leonhardi, Huang, Loyd, Paola, Paranjape, Liu, Wu, Ni, Hancock, Wasti, Spence, Stojkovic, Gamido, Montalvo, Parker, Burton, Mejia, Liu, Wang, Kim, Zhou, Hu, Chu, Cai, Tindal, Feichtenhofer, Gao, Civin, Beaty, Kreymer, Li, Adkins, Xu, Testuggine, David, Parikh, Liskovich, Foss, Wang, Le, Holland, Dowling, Jamil, Montgomery, Presani, Hahn, Wood, Le, Brinkman, Arcaute, Dunbar, Smothers, Sun, Kreuk, Tian, Kokkinos, Ozgenel, Caggioni, Kanayet, Seide, Florez, Schwarz, Badeer, Swee, Halpern, Herman, Sizov, Guangyi, Zhang, Lakshminarayanan, Inan, Shojanazeri, Zou, Wang, Zha, Habeeb, Rudolph, Suk, Aspegren, Goldman, Zhan, Damlaj, Molybog, Tufanov, Leontiadis, Veliche, Gat, Weissman, Geboski, Kohli, Lam, Asher, Gaya, Marcus, Tang, Chan, Zhen, Reizenstein, Teboul, Zhong, Jin, Yang, Cummings, Carvill, Shepard, McPhie,
  Torres, Ginsburg, Wang, Wu, U, Saxena, Khandelwal, Zand, Matosich, Veeraraghavan, Michelena, Li, Jagadeesh, Huang, Chawla, Huang, Chen, Garg, A, Silva, Bell, Zhang, Guo, Yu, Moshkovich, Wehrstedt, Khabsa, Avalani, Bhatt, Mankus, Hasson, Lennie, Reso, Groshev, Naumov, Lathi, Keneally, Liu, Seltzer, Valko, Restrepo, Patel, Vyatskov, Samvelyan, Clark, Macey, Wang, Hermoso, Metanat, Rastegari, Bansal, Santhanam, Parks, White, Bawa, Singhal, Egebo, Usunier, Mehta, Laptev, Dong, Cheng, Chernoguz, Hart, Salpekar, Kalinli, Kent, Parekh, Saab, Balaji, Rittner, Bontrager, Roux, Dollar, Zvyagina, Ratanchandani, Yuvraj, Liang, Alao, Rodriguez, Ayub, Murthy, Nayani, Mitra, Parthasarathy, Li, Hogan, Battey, Wang, Howes, Rinott, Mehta, Siby, Bondu, Datta, Chugh, Hunt, Dhillon, Sidorov, Pan, Mahajan, Verma, Yamamoto, Ramaswamy, Lindsay, Lindsay, Feng, Lin, Zha, Patil, Shankar, Zhang, Zhang, Wang, Agarwal, Sajuyigbe, Chintala, Max, Chen, Kehoe, Satterfield, Govindaprasad, Gupta, Deng, Cho, Virk, Subramanian, Choudhury,
  Goldman, Remez, Glaser, Best, Koehler, Robinson, Li, Zhang, Matthews, Chou, Shaked, Vontimitta, Ajayi, Montanez, Mohan, Kumar, Mangla, Ionescu, Poenaru, Mihailescu, Ivanov, Li, Wang, Jiang, Bouaziz, Constable, Tang, Wu, Wang, Wu, Gao, Kleinman, Chen, Hu, Jia, Qi, Li, Zhang, Zhang, Adi, Nam, Yu, Wang, Zhao, Hao, Qian, Li, He, Rait, DeVito, Rosnbrick, Wen, Yang, Zhao, and Ma]{llama3}
Aaron Grattafiori, Abhimanyu Dubey, Abhinav Jauhri, Abhinav Pandey, Abhishek Kadian, Ahmad Al-Dahle, Aiesha Letman, Akhil Mathur, Alan Schelten, Alex Vaughan, Amy Yang, Angela Fan, Anirudh Goyal, Anthony Hartshorn, Aobo Yang, Archi Mitra, Archie Sravankumar, Artem Korenev, Arthur Hinsvark, Arun Rao, Aston Zhang, Aurelien Rodriguez, Austen Gregerson, Ava Spataru, Baptiste Roziere, Bethany Biron, Binh Tang, Bobbie Chern, Charlotte Caucheteux, Chaya Nayak, Chloe Bi, Chris Marra, Chris McConnell, Christian Keller, Christophe Touret, Chunyang Wu, Corinne Wong, Cristian~Canton Ferrer, Cyrus Nikolaidis, Damien Allonsius, Daniel Song, Danielle Pintz, Danny Livshits, Danny Wyatt, David Esiobu, Dhruv Choudhary, Dhruv Mahajan, Diego Garcia-Olano, Diego Perino, Dieuwke Hupkes, Egor Lakomkin, Ehab AlBadawy, Elina Lobanova, Emily Dinan, Eric~Michael Smith, Filip Radenovic, Francisco Guzmán, Frank Zhang, Gabriel Synnaeve, Gabrielle Lee, Georgia~Lewis Anderson, Govind Thattai, Graeme Nail, Gregoire Mialon, Guan Pang,
  Guillem Cucurell, Hailey Nguyen, Hannah Korevaar, Hu Xu, Hugo Touvron, Iliyan Zarov, Imanol~Arrieta Ibarra, Isabel Kloumann, Ishan Misra, Ivan Evtimov, Jack Zhang, Jade Copet, Jaewon Lee, Jan Geffert, Jana Vranes, Jason Park, Jay Mahadeokar, Jeet Shah, Jelmer van~der Linde, Jennifer Billock, Jenny Hong, Jenya Lee, Jeremy Fu, Jianfeng Chi, Jianyu Huang, Jiawen Liu, Jie Wang, Jiecao Yu, Joanna Bitton, Joe Spisak, Jongsoo Park, Joseph Rocca, Joshua Johnstun, Joshua Saxe, Junteng Jia, Kalyan~Vasuden Alwala, Karthik Prasad, Kartikeya Upasani, Kate Plawiak, Ke Li, Kenneth Heafield, Kevin Stone, Khalid El-Arini, Krithika Iyer, Kshitiz Malik, Kuenley Chiu, Kunal Bhalla, Kushal Lakhotia, Lauren Rantala-Yeary, Laurens van~der Maaten, Lawrence Chen, Liang Tan, Liz Jenkins, Louis Martin, Lovish Madaan, Lubo Malo, Lukas Blecher, Lukas Landzaat, Luke de Oliveira, Madeline Muzzi, Mahesh Pasupuleti, Mannat Singh, Manohar Paluri, Marcin Kardas, Maria Tsimpoukelli, Mathew Oldham, Mathieu Rita, Maya Pavlova, Melanie Kambadur,
  Mike Lewis, Min Si, Mitesh~Kumar Singh, Mona Hassan, Naman Goyal, Narjes Torabi, Nikolay Bashlykov, Nikolay Bogoychev, Niladri Chatterji, Ning Zhang, Olivier Duchenne, Onur Çelebi, Patrick Alrassy, Pengchuan Zhang, Pengwei Li, Petar Vasic, Peter Weng, Prajjwal Bhargava, Pratik Dubal, Praveen Krishnan, Punit~Singh Koura, Puxin Xu, Qing He, Qingxiao Dong, Ragavan Srinivasan, Raj Ganapathy, Ramon Calderer, Ricardo~Silveira Cabral, Robert Stojnic, Roberta Raileanu, Rohan Maheswari, Rohit Girdhar, Rohit Patel, Romain Sauvestre, Ronnie Polidoro, Roshan Sumbaly, Ross Taylor, Ruan Silva, Rui Hou, Rui Wang, Saghar Hosseini, Sahana Chennabasappa, Sanjay Singh, Sean Bell, Seohyun~Sonia Kim, Sergey Edunov, Shaoliang Nie, Sharan Narang, Sharath Raparthy, Sheng Shen, Shengye Wan, Shruti Bhosale, Shun Zhang, Simon Vandenhende, Soumya Batra, Spencer Whitman, Sten Sootla, Stephane Collot, Suchin Gururangan, Sydney Borodinsky, Tamar Herman, Tara Fowler, Tarek Sheasha, Thomas Georgiou, Thomas Scialom, Tobias Speckbacher,
  Todor Mihaylov, Tong Xiao, Ujjwal Karn, Vedanuj Goswami, Vibhor Gupta, Vignesh Ramanathan, Viktor Kerkez, Vincent Gonguet, Virginie Do, Vish Vogeti, Vítor Albiero, Vladan Petrovic, Weiwei Chu, Wenhan Xiong, Wenyin Fu, Whitney Meers, Xavier Martinet, Xiaodong Wang, Xiaofang Wang, Xiaoqing~Ellen Tan, Xide Xia, Xinfeng Xie, Xuchao Jia, Xuewei Wang, Yaelle Goldschlag, Yashesh Gaur, Yasmine Babaei, Yi Wen, Yiwen Song, Yuchen Zhang, Yue Li, Yuning Mao, Zacharie~Delpierre Coudert, Zheng Yan, Zhengxing Chen, Zoe Papakipos, Aaditya Singh, Aayushi Srivastava, Abha Jain, Adam Kelsey, Adam Shajnfeld, Adithya Gangidi, Adolfo Victoria, Ahuva Goldstand, Ajay Menon, Ajay Sharma, Alex Boesenberg, Alexei Baevski, Allie Feinstein, Amanda Kallet, Amit Sangani, Amos Teo, Anam Yunus, Andrei Lupu, Andres Alvarado, Andrew Caples, Andrew Gu, Andrew Ho, Andrew Poulton, Andrew Ryan, Ankit Ramchandani, Annie Dong, Annie Franco, Anuj Goyal, Aparajita Saraf, Arkabandhu Chowdhury, Ashley Gabriel, Ashwin Bharambe, Assaf Eisenman, Azadeh
  Yazdan, Beau James, Ben Maurer, Benjamin Leonhardi, Bernie Huang, Beth Loyd, Beto~De Paola, Bhargavi Paranjape, Bing Liu, Bo Wu, Boyu Ni, Braden Hancock, Bram Wasti, Brandon Spence, Brani Stojkovic, Brian Gamido, Britt Montalvo, Carl Parker, Carly Burton, Catalina Mejia, Ce Liu, Changhan Wang, Changkyu Kim, Chao Zhou, Chester Hu, Ching-Hsiang Chu, Chris Cai, Chris Tindal, Christoph Feichtenhofer, Cynthia Gao, Damon Civin, Dana Beaty, Daniel Kreymer, Daniel Li, David Adkins, David Xu, Davide Testuggine, Delia David, Devi Parikh, Diana Liskovich, Didem Foss, Dingkang Wang, Duc Le, Dustin Holland, Edward Dowling, Eissa Jamil, Elaine Montgomery, Eleonora Presani, Emily Hahn, Emily Wood, Eric-Tuan Le, Erik Brinkman, Esteban Arcaute, Evan Dunbar, Evan Smothers, Fei Sun, Felix Kreuk, Feng Tian, Filippos Kokkinos, Firat Ozgenel, Francesco Caggioni, Frank Kanayet, Frank Seide, Gabriela~Medina Florez, Gabriella Schwarz, Gada Badeer, Georgia Swee, Gil Halpern, Grant Herman, Grigory Sizov, Guangyi, Zhang, Guna
  Lakshminarayanan, Hakan Inan, Hamid Shojanazeri, Han Zou, Hannah Wang, Hanwen Zha, Haroun Habeeb, Harrison Rudolph, Helen Suk, Henry Aspegren, Hunter Goldman, Hongyuan Zhan, Ibrahim Damlaj, Igor Molybog, Igor Tufanov, Ilias Leontiadis, Irina-Elena Veliche, Itai Gat, Jake Weissman, James Geboski, James Kohli, Janice Lam, Japhet Asher, Jean-Baptiste Gaya, Jeff Marcus, Jeff Tang, Jennifer Chan, Jenny Zhen, Jeremy Reizenstein, Jeremy Teboul, Jessica Zhong, Jian Jin, Jingyi Yang, Joe Cummings, Jon Carvill, Jon Shepard, Jonathan McPhie, Jonathan Torres, Josh Ginsburg, Junjie Wang, Kai Wu, Kam~Hou U, Karan Saxena, Kartikay Khandelwal, Katayoun Zand, Kathy Matosich, Kaushik Veeraraghavan, Kelly Michelena, Keqian Li, Kiran Jagadeesh, Kun Huang, Kunal Chawla, Kyle Huang, Lailin Chen, Lakshya Garg, Lavender A, Leandro Silva, Lee Bell, Lei Zhang, Liangpeng Guo, Licheng Yu, Liron Moshkovich, Luca Wehrstedt, Madian Khabsa, Manav Avalani, Manish Bhatt, Martynas Mankus, Matan Hasson, Matthew Lennie, Matthias Reso, Maxim
  Groshev, Maxim Naumov, Maya Lathi, Meghan Keneally, Miao Liu, Michael~L. Seltzer, Michal Valko, Michelle Restrepo, Mihir Patel, Mik Vyatskov, Mikayel Samvelyan, Mike Clark, Mike Macey, Mike Wang, Miquel~Jubert Hermoso, Mo Metanat, Mohammad Rastegari, Munish Bansal, Nandhini Santhanam, Natascha Parks, Natasha White, Navyata Bawa, Nayan Singhal, Nick Egebo, Nicolas Usunier, Nikhil Mehta, Nikolay~Pavlovich Laptev, Ning Dong, Norman Cheng, Oleg Chernoguz, Olivia Hart, Omkar Salpekar, Ozlem Kalinli, Parkin Kent, Parth Parekh, Paul Saab, Pavan Balaji, Pedro Rittner, Philip Bontrager, Pierre Roux, Piotr Dollar, Polina Zvyagina, Prashant Ratanchandani, Pritish Yuvraj, Qian Liang, Rachad Alao, Rachel Rodriguez, Rafi Ayub, Raghotham Murthy, Raghu Nayani, Rahul Mitra, Rangaprabhu Parthasarathy, Raymond Li, Rebekkah Hogan, Robin Battey, Rocky Wang, Russ Howes, Ruty Rinott, Sachin Mehta, Sachin Siby, Sai~Jayesh Bondu, Samyak Datta, Sara Chugh, Sara Hunt, Sargun Dhillon, Sasha Sidorov, Satadru Pan, Saurabh Mahajan,
  Saurabh Verma, Seiji Yamamoto, Sharadh Ramaswamy, Shaun Lindsay, Shaun Lindsay, Sheng Feng, Shenghao Lin, Shengxin~Cindy Zha, Shishir Patil, Shiva Shankar, Shuqiang Zhang, Shuqiang Zhang, Sinong Wang, Sneha Agarwal, Soji Sajuyigbe, Soumith Chintala, Stephanie Max, Stephen Chen, Steve Kehoe, Steve Satterfield, Sudarshan Govindaprasad, Sumit Gupta, Summer Deng, Sungmin Cho, Sunny Virk, Suraj Subramanian, Sy Choudhury, Sydney Goldman, Tal Remez, Tamar Glaser, Tamara Best, Thilo Koehler, Thomas Robinson, Tianhe Li, Tianjun Zhang, Tim Matthews, Timothy Chou, Tzook Shaked, Varun Vontimitta, Victoria Ajayi, Victoria Montanez, Vijai Mohan, Vinay~Satish Kumar, Vishal Mangla, Vlad Ionescu, Vlad Poenaru, Vlad~Tiberiu Mihailescu, Vladimir Ivanov, Wei Li, Wenchen Wang, Wenwen Jiang, Wes Bouaziz, Will Constable, Xiaocheng Tang, Xiaojian Wu, Xiaolan Wang, Xilun Wu, Xinbo Gao, Yaniv Kleinman, Yanjun Chen, Ye Hu, Ye Jia, Ye Qi, Yenda Li, Yilin Zhang, Ying Zhang, Yossi Adi, Youngjin Nam, Yu, Wang, Yu Zhao, Yuchen Hao, Yundi
  Qian, Yunlu Li, Yuzi He, Zach Rait, Zachary DeVito, Zef Rosnbrick, Zhaoduo Wen, Zhenyu Yang, Zhiwei Zhao, and Zhiyu Ma.
\newblock The llama 3 herd of models, 2024.

\bibitem[Hall et~al.(2022)Hall, van~der Maaten, Gustafson, and Adcock]{Hall2022ASS}
Melissa Hall, Laurens van~der Maaten, Laura Gustafson, and Aaron~B. Adcock.
\newblock A systematic study of bias amplification.
\newblock \emph{ArXiv}, abs/2201.11706, 2022.

\bibitem[Hassan et~al.(2021)Hassan, Huenerfauth, and Alm]{hassan2021unpacking-ableist-intersectional}
Saad Hassan, Matt Huenerfauth, and Cecilia~Ovesdotter Alm.
\newblock Unpacking the interdependent systems of discrimination: Ableist bias in nlp systems through an intersectional lens.
\newblock \emph{arXiv preprint arXiv:2110.00521}, 2021.

\bibitem[Kamruzzaman and Kim(2025)]{impact-disability-fairness-bias}
Mahammed Kamruzzaman and Gene~Louis Kim.
\newblock The impact of disability disclosure on fairness and bias in llm-driven candidate selection.
\newblock \emph{The International FLAIRS Conference Proceedings}, 2025.

\bibitem[Kamruzzaman et~al.(2024)Kamruzzaman, Shovon, and Kim]{kamruzzaman-etal-2024-investigating}
Mahammed Kamruzzaman, Md. Shovon, and Gene Kim.
\newblock Investigating subtler biases in {LLM}s: Ageism, beauty, institutional, and nationality bias in generative models.
\newblock In \emph{Findings of the Association for Computational Linguistics: ACL 2024}, pages 8940--8965, Bangkok, Thailand, 2024. Association for Computational Linguistics.

\bibitem[Kotek et~al.(2023)Kotek, Dockum, and Sun]{Kotek2023GenderBA}
Hadas Kotek, Rikker Dockum, and David~Q. Sun.
\newblock Gender bias and stereotypes in large language models.
\newblock \emph{Proceedings of The ACM Collective Intelligence Conference}, 2023.

\bibitem[Kumar et~al.(2022)Kumar, Balachandran, Njoo, Anastasopoulos, and Tsvetkov]{Kumar2022LanguageGM}
Sachin Kumar, Vidhisha Balachandran, Lucille Njoo, Antonios Anastasopoulos, and Yulia Tsvetkov.
\newblock Language generation models can cause harm: So what can we do about it? an actionable survey.
\newblock In \emph{Conference of the European Chapter of the Association for Computational Linguistics}, 2022.

\bibitem[Li et~al.(2024)Li, Kamaraj, Ma, and Ebling]{decoding-ableism-intersectional-llm}
Rong Li, Ashwini Kamaraj, Jing Ma, and Sarah Ebling.
\newblock Decoding ableism in large language models: An intersectional approach.
\newblock \emph{Proceedings of the Third Workshop on NLP for Positive Impact}, 2024.

\bibitem[Ma et~al.(2023)Ma, Chiang, Wu, Wang, and Vosoughi]{intersectional-stereotypes}
Weicheng Ma, Brian Chiang, Tong Wu, Lili Wang, and Soroush Vosoughi.
\newblock Intersectional stereotypes in large language models: Dataset and analysis.
\newblock In \emph{Findings of the Association for Computational Linguistics: EMNLP 2023}, pages 8589--8597, Singapore, 2023. Association for Computational Linguistics.

\bibitem[Madhusudhan et~al.(2024)Madhusudhan, Madhusudhan, Yadav, and Hashemi]{madhusudhan2024llmsknowanswerinvestigating}
Nishanth Madhusudhan, Sathwik~Tejaswi Madhusudhan, Vikas Yadav, and Masoud Hashemi.
\newblock Do llms know when to not answer? investigating abstention abilities of large language models, 2024.

\bibitem[Meghwani et~al.(2025)Meghwani, Agarwal, Pattnayak, Patel, and Panda]{meghwani-etal-2025-hard}
Hansa Meghwani, Amit Agarwal, Priyaranjan Pattnayak, Hitesh~Laxmichand Patel, and Srikant Panda.
\newblock Hard negative mining for domain-specific retrieval in enterprise systems.
\newblock In \emph{Proceedings of the 63rd Annual Meeting of the Association for Computational Linguistics (Volume 6: Industry Track)}, pages 1013--1026, Vienna, Austria, 2025. Association for Computational Linguistics.

\bibitem[Microsoft et~al.(2025)Microsoft, :, Abouelenin, Ashfaq, Atkinson, Awadalla, Bach, Bao, Benhaim, Cai, Chaudhary, Chen, Chen, Chen, Chen, Chen, Chen, ling Chen, Dai, Dai, Fan, Gao, Gao, Garg, Goswami, Hao, Hendy, Hu, Jin, Khademi, Kim, Kim, Lee, Li, Li, Liang, Lin, Lin, Liu, Liu, Lopez, Luo, Madan, Mazalov, Mitra, Mousavi, Nguyen, Pan, Perez-Becker, Platin, Portet, Qiu, Ren, Ren, Roy, Shang, Shen, Singhal, Som, Song, Sych, Vaddamanu, Wang, Wang, Wang, Wu, Xu, Xu, Yang, Yang, Yu, Zabir, Zhang, Zhang, Zhang, and Zhou]{phi4mini}
Microsoft, :, Abdelrahman Abouelenin, Atabak Ashfaq, Adam Atkinson, Hany Awadalla, Nguyen Bach, Jianmin Bao, Alon Benhaim, Martin Cai, Vishrav Chaudhary, Congcong Chen, Dong Chen, Dongdong Chen, Junkun Chen, Weizhu Chen, Yen-Chun Chen, Yi ling Chen, Qi Dai, Xiyang Dai, Ruchao Fan, Mei Gao, Min Gao, Amit Garg, Abhishek Goswami, Junheng Hao, Amr Hendy, Yuxuan Hu, Xin Jin, Mahmoud Khademi, Dongwoo Kim, Young~Jin Kim, Gina Lee, Jinyu Li, Yunsheng Li, Chen Liang, Xihui Lin, Zeqi Lin, Mengchen Liu, Yang Liu, Gilsinia Lopez, Chong Luo, Piyush Madan, Vadim Mazalov, Arindam Mitra, Ali Mousavi, Anh Nguyen, Jing Pan, Daniel Perez-Becker, Jacob Platin, Thomas Portet, Kai Qiu, Bo Ren, Liliang Ren, Sambuddha Roy, Ning Shang, Yelong Shen, Saksham Singhal, Subhojit Som, Xia Song, Tetyana Sych, Praneetha Vaddamanu, Shuohang Wang, Yiming Wang, Zhenghao Wang, Haibin Wu, Haoran Xu, Weijian Xu, Yifan Yang, Ziyi Yang, Donghan Yu, Ishmam Zabir, Jianwen Zhang, Li~Lyna Zhang, Yunan Zhang, and Xiren Zhou.
\newblock Phi-4-mini technical report: Compact yet powerful multimodal language models via mixture-of-loras, 2025.

\bibitem[Nadeem et~al.(2020)Nadeem, Bethke, and Reddy]{Nadeem2020StereoSetMS}
Moin Nadeem, Anna Bethke, and Siva Reddy.
\newblock Stereoset: Measuring stereotypical bias in pretrained language models.
\newblock In \emph{Annual Meeting of the Association for Computational Linguistics}, 2020.

\bibitem[OpenAI et~al.(2024)OpenAI, Achiam, Adler, Agarwal, Ahmad, Akkaya, Aleman, Almeida, Altenschmidt, Altman, Anadkat, Avila, Babuschkin, Balaji, Balcom, Baltescu, Bao, Bavarian, Belgum, Bello, Berdine, Bernadett-Shapiro, Berner, Bogdonoff, Boiko, Boyd, Brakman, Brockman, Brooks, Brundage, Button, Cai, Campbell, Cann, Carey, Carlson, Carmichael, Chan, Chang, Chantzis, Chen, Chen, Chen, Chen, Chen, Chess, Cho, Chu, Chung, Cummings, Currier, Dai, Decareaux, Degry, Deutsch, Deville, Dhar, Dohan, Dowling, Dunning, Ecoffet, Eleti, Eloundou, Farhi, Fedus, Felix, Fishman, Forte, Fulford, Gao, Georges, Gibson, Goel, Gogineni, Goh, Gontijo-Lopes, Gordon, Grafstein, Gray, Greene, Gross, Gu, Guo, Hallacy, Han, Harris, He, Heaton, Heidecke, Hesse, Hickey, Hickey, Hoeschele, Houghton, Hsu, Hu, Hu, Huizinga, Jain, Jain, Jang, Jiang, Jiang, Jin, Jin, Jomoto, Jonn, Jun, Kaftan, Łukasz Kaiser, Kamali, Kanitscheider, Keskar, Khan, Kilpatrick, Kim, Kim, Kim, Kirchner, Kiros, Knight, Kokotajlo, Łukasz Kondraciuk, Kondrich,
  Konstantinidis, Kosic, Krueger, Kuo, Lampe, Lan, Lee, Leike, Leung, Levy, Li, Lim, Lin, Lin, Litwin, Lopez, Lowe, Lue, Makanju, Malfacini, Manning, Markov, Markovski, Martin, Mayer, Mayne, McGrew, McKinney, McLeavey, McMillan, McNeil, Medina, Mehta, Menick, Metz, Mishchenko, Mishkin, Monaco, Morikawa, Mossing, Mu, Murati, Murk, Mély, Nair, Nakano, Nayak, Neelakantan, Ngo, Noh, Ouyang, O'Keefe, Pachocki, Paino, Palermo, Pantuliano, Parascandolo, Parish, Parparita, Passos, Pavlov, Peng, Perelman, de~Avila Belbute~Peres, Petrov, de~Oliveira~Pinto, Michael, Pokorny, Pokrass, Pong, Powell, Power, Power, Proehl, Puri, Radford, Rae, Ramesh, Raymond, Real, Rimbach, Ross, Rotsted, Roussez, Ryder, Saltarelli, Sanders, Santurkar, Sastry, Schmidt, Schnurr, Schulman, Selsam, Sheppard, Sherbakov, Shieh, Shoker, Shyam, Sidor, Sigler, Simens, Sitkin, Slama, Sohl, Sokolowsky, Song, Staudacher, Such, Summers, Sutskever, Tang, Tezak, Thompson, Tillet, Tootoonchian, Tseng, Tuggle, Turley, Tworek, Uribe, Vallone, Vijayvergiya,
  Voss, Wainwright, Wang, Wang, Wang, Ward, Wei, Weinmann, Welihinda, Welinder, Weng, Weng, Wiethoff, Willner, Winter, Wolrich, Wong, Workman, Wu, Wu, Wu, Xiao, Xu, Yoo, Yu, Yuan, Zaremba, Zellers, Zhang, Zhang, Zhao, Zheng, Zhuang, Zhuk, and Zoph]{openai2024gpt4technicalreport}
OpenAI, Josh Achiam, Steven Adler, Sandhini Agarwal, Lama Ahmad, Ilge Akkaya, Florencia~Leoni Aleman, Diogo Almeida, Janko Altenschmidt, Sam Altman, Shyamal Anadkat, Red Avila, Igor Babuschkin, Suchir Balaji, Valerie Balcom, Paul Baltescu, Haiming Bao, Mohammad Bavarian, Jeff Belgum, Irwan Bello, Jake Berdine, Gabriel Bernadett-Shapiro, Christopher Berner, Lenny Bogdonoff, Oleg Boiko, Madelaine Boyd, Anna-Luisa Brakman, Greg Brockman, Tim Brooks, Miles Brundage, Kevin Button, Trevor Cai, Rosie Campbell, Andrew Cann, Brittany Carey, Chelsea Carlson, Rory Carmichael, Brooke Chan, Che Chang, Fotis Chantzis, Derek Chen, Sully Chen, Ruby Chen, Jason Chen, Mark Chen, Ben Chess, Chester Cho, Casey Chu, Hyung~Won Chung, Dave Cummings, Jeremiah Currier, Yunxing Dai, Cory Decareaux, Thomas Degry, Noah Deutsch, Damien Deville, Arka Dhar, David Dohan, Steve Dowling, Sheila Dunning, Adrien Ecoffet, Atty Eleti, Tyna Eloundou, David Farhi, Liam Fedus, Niko Felix, Simón~Posada Fishman, Juston Forte, Isabella Fulford, Leo
  Gao, Elie Georges, Christian Gibson, Vik Goel, Tarun Gogineni, Gabriel Goh, Rapha Gontijo-Lopes, Jonathan Gordon, Morgan Grafstein, Scott Gray, Ryan Greene, Joshua Gross, Shixiang~Shane Gu, Yufei Guo, Chris Hallacy, Jesse Han, Jeff Harris, Yuchen He, Mike Heaton, Johannes Heidecke, Chris Hesse, Alan Hickey, Wade Hickey, Peter Hoeschele, Brandon Houghton, Kenny Hsu, Shengli Hu, Xin Hu, Joost Huizinga, Shantanu Jain, Shawn Jain, Joanne Jang, Angela Jiang, Roger Jiang, Haozhun Jin, Denny Jin, Shino Jomoto, Billie Jonn, Heewoo Jun, Tomer Kaftan, Łukasz Kaiser, Ali Kamali, Ingmar Kanitscheider, Nitish~Shirish Keskar, Tabarak Khan, Logan Kilpatrick, Jong~Wook Kim, Christina Kim, Yongjik Kim, Jan~Hendrik Kirchner, Jamie Kiros, Matt Knight, Daniel Kokotajlo, Łukasz Kondraciuk, Andrew Kondrich, Aris Konstantinidis, Kyle Kosic, Gretchen Krueger, Vishal Kuo, Michael Lampe, Ikai Lan, Teddy Lee, Jan Leike, Jade Leung, Daniel Levy, Chak~Ming Li, Rachel Lim, Molly Lin, Stephanie Lin, Mateusz Litwin, Theresa Lopez, Ryan
  Lowe, Patricia Lue, Anna Makanju, Kim Malfacini, Sam Manning, Todor Markov, Yaniv Markovski, Bianca Martin, Katie Mayer, Andrew Mayne, Bob McGrew, Scott~Mayer McKinney, Christine McLeavey, Paul McMillan, Jake McNeil, David Medina, Aalok Mehta, Jacob Menick, Luke Metz, Andrey Mishchenko, Pamela Mishkin, Vinnie Monaco, Evan Morikawa, Daniel Mossing, Tong Mu, Mira Murati, Oleg Murk, David Mély, Ashvin Nair, Reiichiro Nakano, Rajeev Nayak, Arvind Neelakantan, Richard Ngo, Hyeonwoo Noh, Long Ouyang, Cullen O'Keefe, Jakub Pachocki, Alex Paino, Joe Palermo, Ashley Pantuliano, Giambattista Parascandolo, Joel Parish, Emy Parparita, Alex Passos, Mikhail Pavlov, Andrew Peng, Adam Perelman, Filipe de Avila Belbute~Peres, Michael Petrov, Henrique~Ponde de Oliveira~Pinto, Michael, Pokorny, Michelle Pokrass, Vitchyr~H. Pong, Tolly Powell, Alethea Power, Boris Power, Elizabeth Proehl, Raul Puri, Alec Radford, Jack Rae, Aditya Ramesh, Cameron Raymond, Francis Real, Kendra Rimbach, Carl Ross, Bob Rotsted, Henri Roussez,
  Nick Ryder, Mario Saltarelli, Ted Sanders, Shibani Santurkar, Girish Sastry, Heather Schmidt, David Schnurr, John Schulman, Daniel Selsam, Kyla Sheppard, Toki Sherbakov, Jessica Shieh, Sarah Shoker, Pranav Shyam, Szymon Sidor, Eric Sigler, Maddie Simens, Jordan Sitkin, Katarina Slama, Ian Sohl, Benjamin Sokolowsky, Yang Song, Natalie Staudacher, Felipe~Petroski Such, Natalie Summers, Ilya Sutskever, Jie Tang, Nikolas Tezak, Madeleine~B. Thompson, Phil Tillet, Amin Tootoonchian, Elizabeth Tseng, Preston Tuggle, Nick Turley, Jerry Tworek, Juan Felipe~Cerón Uribe, Andrea Vallone, Arun Vijayvergiya, Chelsea Voss, Carroll Wainwright, Justin~Jay Wang, Alvin Wang, Ben Wang, Jonathan Ward, Jason Wei, CJ Weinmann, Akila Welihinda, Peter Welinder, Jiayi Weng, Lilian Weng, Matt Wiethoff, Dave Willner, Clemens Winter, Samuel Wolrich, Hannah Wong, Lauren Workman, Sherwin Wu, Jeff Wu, Michael Wu, Kai Xiao, Tao Xu, Sarah Yoo, Kevin Yu, Qiming Yuan, Wojciech Zaremba, Rowan Zellers, Chong Zhang, Marvin Zhang, Shengjia
  Zhao, Tianhao Zheng, Juntang Zhuang, William Zhuk, and Barret Zoph.
\newblock Gpt-4 technical report, 2024.

\bibitem[Panda et~al.(2025{\natexlab{a}})Panda, Agarwal, and Patel]{AccessEvalDataset}
Srikant Panda, Amit Agarwal, and Hitesh~Laxmichand Patel.
\newblock Accesseval, 2025{\natexlab{a}}.

\bibitem[Panda et~al.(2025{\natexlab{b}})Panda, Agarwal, and Patel]{panda2025accessevalbenchmarkingdisabilitybias}
Srikant Panda, Amit Agarwal, and Hitesh~Laxmichand Patel.
\newblock Accesseval: Benchmarking disability bias in large language models, 2025{\natexlab{b}}.

\bibitem[Panda et~al.(2025{\natexlab{c}})Panda, Patel, Al-Khalifa, Agarwal, Al-Khalifa, and Al-Ghamdi]{panda2025daiqauditingdemographicattribute}
Srikant Panda, Hitesh~Laxmichand Patel, Shahad Al-Khalifa, Amit Agarwal, Hend Al-Khalifa, and Sharefah Al-Ghamdi.
\newblock Daiq: Auditing demographic attribute inference from question in llms, 2025{\natexlab{c}}.

\bibitem[Patel et~al.(2025{\natexlab{a}})Patel, Agarwal, Das, Kumar, Panda, Pattnayak, Rafi, Kumar, and Chae]{patel2025sweeval}
Hitesh~Laxmichand Patel, Amit Agarwal, Arion Das, Bhargava Kumar, Srikant Panda, Priyaranjan Pattnayak, Taki~Hasan Rafi, Tejaswini Kumar, and Dong-Kyu Chae.
\newblock Sweeval: Do llms really swear? a safety benchmark for testing limits for enterprise use.
\newblock In \emph{Proceedings of the 2025 Conference of the Nations of the Americas Chapter of the Association for Computational Linguistics: Human Language Technologies (Volume 3: Industry Track)}, pages 558--582, 2025{\natexlab{a}}.

\bibitem[Patel et~al.(2025{\natexlab{b}})Patel, Agarwal, Panda, Meghwani, Dua, Li, Sheng, Ravi, and Roth]{patel2025pcrimeasuringcontextrobustness}
Hitesh~Laxmichand Patel, Amit Agarwal, Srikant Panda, Hansa Meghwani, Karan Dua, Paul Li, Tao Sheng, Sujith Ravi, and Dan Roth.
\newblock Pcri: Measuring context robustness in multimodal models for enterprise applications, 2025{\natexlab{b}}.

\bibitem[Pattnayak et~al.(2025)Pattnayak, Agarwal, Meghwani, Patel, and Panda]{pattnayak2025hybrid}
Priyaranjan Pattnayak, Amit Agarwal, Hansa Meghwani, Hitesh~Laxmichand Patel, and Srikant Panda.
\newblock Hybrid ai for responsive multi-turn online conversations with novel dynamic routing and feedback adaptation.
\newblock In \emph{Proceedings of the 4th International Workshop on Knowledge-Augmented Methods for Natural Language Processing}, pages 215--229, 2025.

\bibitem[Qwen et~al.(2025)Qwen, Yang, Yang, Zhang, Hui, Zheng, Yu, Li, Liu, Huang, Wei, Lin, Yang, Tu, Zhang, Yang, Yang, Zhou, Lin, Dang, Lu, Bao, Yang, Yu, Li, Xue, Zhang, Zhu, Men, Lin, Li, Tang, Xia, Ren, Ren, Fan, Su, Zhang, Wan, Liu, Cui, Zhang, and Qiu]{qwen2.5}
Qwen, An Yang, Baosong Yang, Beichen Zhang, Binyuan Hui, Bo Zheng, Bowen Yu, Chengyuan Li, Dayiheng Liu, Fei Huang, Haoran Wei, Huan Lin, Jian Yang, Jianhong Tu, Jianwei Zhang, Jianxin Yang, Jiaxi Yang, Jingren Zhou, Junyang Lin, Kai Dang, Keming Lu, Keqin Bao, Kexin Yang, Le Yu, Mei Li, Mingfeng Xue, Pei Zhang, Qin Zhu, Rui Men, Runji Lin, Tianhao Li, Tianyi Tang, Tingyu Xia, Xingzhang Ren, Xuancheng Ren, Yang Fan, Yang Su, Yichang Zhang, Yu Wan, Yuqiong Liu, Zeyu Cui, Zhenru Zhang, and Zihan Qiu.
\newblock Qwen2.5 technical report, 2025.

\bibitem[Radford and Narasimhan(2018)]{Radford2018ImprovingLU}
Alec Radford and Karthik Narasimhan.
\newblock Improving language understanding by generative pre-training.
\newblock 2018.

\bibitem[Ramtekkar et~al.(2010)Ramtekkar, Reiersen, Todorov, and Todd]{ramtekkar2010sex}
Ujjwal~P Ramtekkar, Angela~M Reiersen, Alexandre~A Todorov, and Richard~D Todd.
\newblock Sex and age differences in attention-deficit/hyperactivity disorder symptoms and diagnoses: implications for dsm-v and icd-11.
\newblock \emph{Journal of the American Academy of Child \& Adolescent Psychiatry}, 49\penalty0 (3):\penalty0 217--228.e1, 2010.

\bibitem[Team(2024)]{ministral}
Mistral~AI Team.
\newblock Un ministral, des ministraux, 2024.

\bibitem[Tolbert and Diana(2023)]{correcting-underrepresentation-intersectional}
Alexander~Williams Tolbert and Emily Diana.
\newblock Correcting underrepresentation and intersectional bias for fair classification.
\newblock \emph{ArXiv}, abs/2306.11112, 2023.

\bibitem[Wan et~al.(2023)Wan, Pu, Sun, Garimella, Chang, and Peng]{wan2023kellywarmpersonjoseph}
Yixin Wan, George Pu, Jiao Sun, Aparna Garimella, Kai-Wei Chang, and Nanyun Peng.
\newblock "kelly is a warm person, joseph is a role model": Gender biases in llm-generated reference letters, 2023.

\bibitem[Wu and Ebling(2024)]{investigating-ableism-mt-conversation}
Guojun Wu and Sarah Ebling.
\newblock Investigating ableism in llms through multi-turn conversation.
\newblock \emph{Proceedings of the Third Workshop on NLP for Positive Impact}, 2024.

\bibitem[Yadkori et~al.(2024)Yadkori, Kuzborskij, Stutz, György, Fisch, Doucet, Beloshapka, Weng, Yang, Szepesvári, Cemgil, and Tomasev]{yadkori2024mitigatingllmhallucinationsconformal}
Yasin~Abbasi Yadkori, Ilja Kuzborskij, David Stutz, András György, Adam Fisch, Arnaud Doucet, Iuliya Beloshapka, Wei-Hung Weng, Yao-Yuan Yang, Csaba Szepesvári, Ali~Taylan Cemgil, and Nenad Tomasev.
\newblock Mitigating llm hallucinations via conformal abstention, 2024.

\bibitem[Yesiltepe et~al.(2024)Yesiltepe, Akdemir, and Yanardag]{mitigate-intersectional-diffusion}
Hidir Yesiltepe, Kiymet Akdemir, and Pinar Yanardag.
\newblock Mist: Mitigating intersectional bias with disentangled cross-attention editing in text-to-image diffusion models.
\newblock \emph{ArXiv}, abs/2403.19738, 2024.

\bibitem[Yuan et~al.(2024)Yuan, Xiong, Zeng, Yu, Jia, Song, and Li]{yuan2024rigorllmresilientguardrailslarge}
Zhuowen Yuan, Zidi Xiong, Yi Zeng, Ning Yu, Ruoxi Jia, Dawn Song, and Bo Li.
\newblock Rigorllm: Resilient guardrails for large language models against undesired content, 2024.

\end{thebibliography}
}

\end{document}